\documentclass[11pt,dvipsnames]{article}

\usepackage[final]{acl}

\usepackage{times}
\usepackage{latexsym}

\usepackage[T1]{fontenc}
\usepackage[utf8]{inputenc}

\usepackage{microtype}

\usepackage{inconsolata}

\usepackage{graphicx}

\usepackage{times}
\usepackage{latexsym}
\usepackage{balance}

\usepackage[T1]{fontenc}

\usepackage[utf8]{inputenc}

\usepackage{microtype}

\usepackage{inconsolata}
\usepackage{graphicx}
\usepackage{amsmath}
\usepackage{amssymb}
\usepackage{graphicx}
\usepackage{enumitem}
\usepackage{array}
\usepackage{booktabs}
\usepackage{colortbl}
\usepackage{amsmath, amssymb, graphicx}
\usepackage[most]{tcolorbox}
\usepackage{amsmath} % For the multline environment
\usepackage{amsmath}   % For math environments and symbols like \mathcal
\usepackage{amsthm}    % For theorem, corollary, and proof environments
\usepackage{float}  % For [H] float placement option
\usepackage{algorithm}
\usepackage{algorithmicx}
\usepackage{algpseudocode}
\usepackage{pifont}
\definecolor{stepblue}{RGB}{48, 129, 208}
\definecolor{commentgreen}{RGB}{60,120,60}
\definecolor{grey}{gray}{0.5}
\definecolor{PineGreen}{RGB}{1,121,111}
\definecolor{PrismBlue}{RGB}{44,123,208}
\definecolor{PrismCite}{RGB}{78,86,165}
\definecolor{PrismLink}{RGB}{35,104,178}
\hypersetup{
    colorlinks=true,
    citecolor=PrismCite,
    linkcolor=PrismLink,
    urlcolor=PrismBlue,
}
\newcommand{\expnum}[1]{\textcolor{PrismBlue}{\raisebox{0.06em}{\footnotesize\ding{#1}}}}
\newcommand{\exppoint}[2]{\noindent\textbf{\expnum{#1}~#2}}

\newcommand{\prismtightspacing}{%
    \setlength{\textfloatsep}{10pt plus 2pt minus 2pt}%
    \setlength{\floatsep}{8pt plus 2pt minus 2pt}%
    \setlength{\intextsep}{8pt plus 2pt minus 2pt}%
    \setlength{\dbltextfloatsep}{10pt plus 2pt minus 2pt}%
    \setlength{\dblfloatsep}{8pt plus 2pt minus 2pt}%
    \setlength{\abovecaptionskip}{4pt plus 1pt minus 1pt}%
    \setlength{\belowcaptionskip}{1pt plus 1pt minus 1pt}%
    \setlength{\abovedisplayskip}{7pt plus 2pt minus 2pt}%
    \setlength{\belowdisplayskip}{7pt plus 2pt minus 2pt}%
    \setlength{\abovedisplayshortskip}{4pt plus 2pt minus 1pt}%
    \setlength{\belowdisplayshortskip}{5pt plus 2pt minus 1pt}%
}

\newtcolorbox{principlebox}[2][]{%
    enhanced,                 % ensure boxed title features work reliably
    breakable,                % safe if content ever spans pages
    colback=PineGreen!5!white,
    colframe=PineGreen!65!black,
    title={#2},
    fonttitle=\bfseries,
    coltitle=black,
    attach boxed title to top left={yshift=-2mm, xshift=3mm},
    boxed title style={%
        colback=PineGreen!12!white,
        colframe=PineGreen!65!black,
    },
    top=1.2mm,
    bottom=1.2mm,
    left=2.5mm,
    right=2.5mm,
    #1,
}

\title{PRISM: Self-Pruning Intrinsic Selection Method for Training-Free Multimodal Data Selection}

\author {
Jinhe Bi\textsuperscript{\rm 1,}\textsuperscript{\rm 2,}\textsuperscript{\rm 5} \quad
Aniri\textsuperscript{\rm 1,}\textsuperscript{\rm 3,}\textsuperscript{\rm 5}\quad
Zengjie Jin\textsuperscript{\rm 1} \quad
Yifan Wang\textsuperscript{\rm 1} \quad
Danqi Yan\textsuperscript{\rm 1}  \quad \\
\textbf{Wenke Huang\textsuperscript{\rm 4}} \quad
\textbf{Xiaowen Ma\textsuperscript{\rm 1}} \quad 
\textbf{Sikuan Yan\textsuperscript{\rm 1}} \quad
\textbf{Artur Hecker\textsuperscript{\rm 2}} \quad
\textbf{Mang Ye\textsuperscript{\rm 4}} \quad \\
\textbf{Xun Xiao\textsuperscript{\rm 2}}\thanks{Corresponding authors: \textit{cognitive.yunpu@gmail.com}, \textit{drxiaoxun@gmail.com}.} \quad  
\textbf{Hinrich Schütze\textsuperscript{\rm 1,}\textsuperscript{\rm 5} }\quad 
\textbf{Volker Tresp\textsuperscript{\rm 1,}\textsuperscript{\rm 5} \quad 
Yunpu Ma\textsuperscript{\rm 1,}\textsuperscript{\rm 5}\footnotemark[1]} \\
\textsuperscript{\rm 1} LMU Munich \quad \textsuperscript{\rm 2} Huawei Heisenberg Research Center \quad \textsuperscript{\rm 3} METEOR\\
\textsuperscript{\rm 4} School of Computer Science, Wuhan University \quad 
\textsuperscript{\rm 5} Munich Center for Machine Learning
}

\begin{document}

\maketitle
\begin{abstract}
Visual instruction tuning adapts pre-trained Multimodal Large Language Models (MLLMs) to follow human instructions for real-world applications. 
However, the rapid growth of these datasets introduces significant redundancy, leading to increased computational costs.
Existing methods for selecting instruction data aim to prune this redundancy, but predominantly rely on computationally demanding techniques such as proxy-based inference or training-based metrics.
Consequently, the substantial computational costs incurred by these selection processes often exacerbate the very efficiency bottlenecks they are intended to resolve, posing a significant challenge to the scalable and effective tuning of MLLMs.
To address this challenge, we first identify a critical, yet previously overlooked, factor: the anisotropy inherent in visual feature distributions. 
We find that this anisotropy induces a \textit{Global Semantic Drift}, and overlooking this phenomenon is a key factor limiting the efficiency of current data selection methods.
Motivated by this insight, we devise \textbf{PRISM}, a training-free framework for efficient visual instruction selection.
PRISM effectively removes the corrupting influence of global background features by modeling the intrinsic visual semantics  via implicit re-centering.
Empirically, PRISM reduces the end-to-end time for data selection and model tuning to just 30\% of conventional pipelines.
More remarkably, it achieves this efficiency while simultaneously enhancing performance, surpassing models fine-tuned on the full dataset across various multimodal and language understanding benchmarks, culminating in a 101.7\% relative performance of the baseline. The code is available for access via  
\href{https://github.com/bibisbar/PRISM_Data_Selection}{this repository}.
\end{abstract}
\begin{figure}[t]
  \includegraphics[width=\columnwidth, trim={4.5cm 6.1cm 8.7cm 2.5cm}, clip]{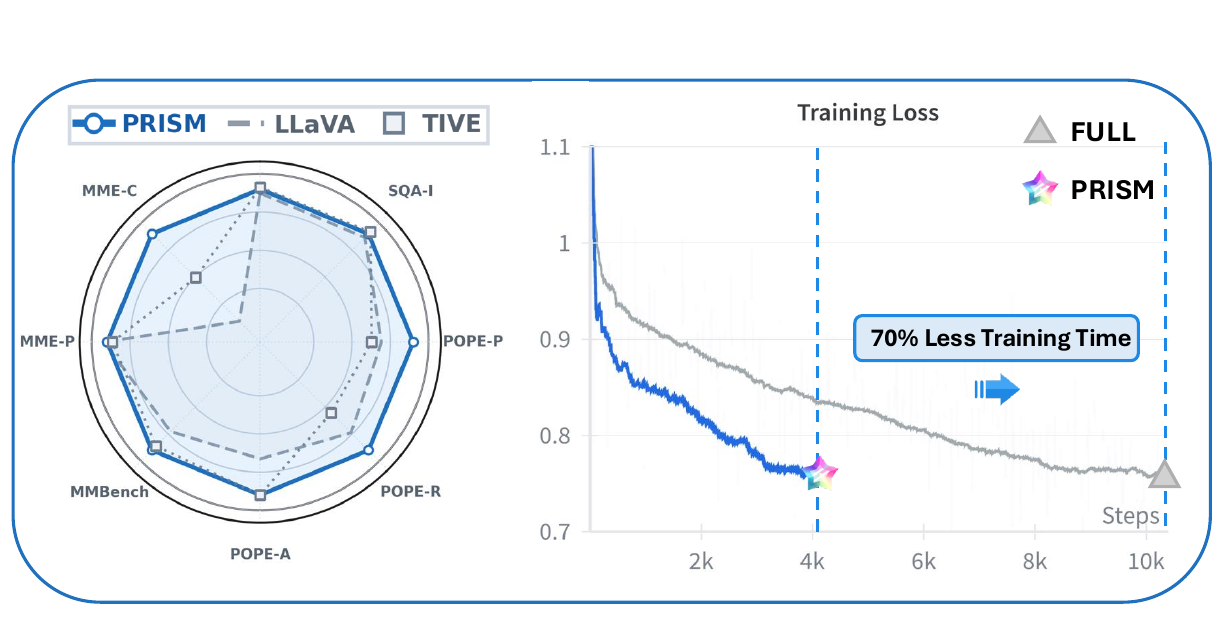}
  \caption{
    \textbf{PRISM achieves state-of-the-art performance with substantially greater training efficiency.} 
    \textbf{(Left)} On a suite of multimodal and language benchmarks, PRISM uniformly outperforms strong baselines like TIVE. 
    \textbf{(Right)} The training loss curve demonstrates that PRISM converges faster, reducing training time by 70\% compared to the LLaVA baseline.
}
  \label{fig:experiments}
\end{figure}
\section{Introduction}

\begin{figure*}[t]
    \centering
    \includegraphics[width=1\linewidth]{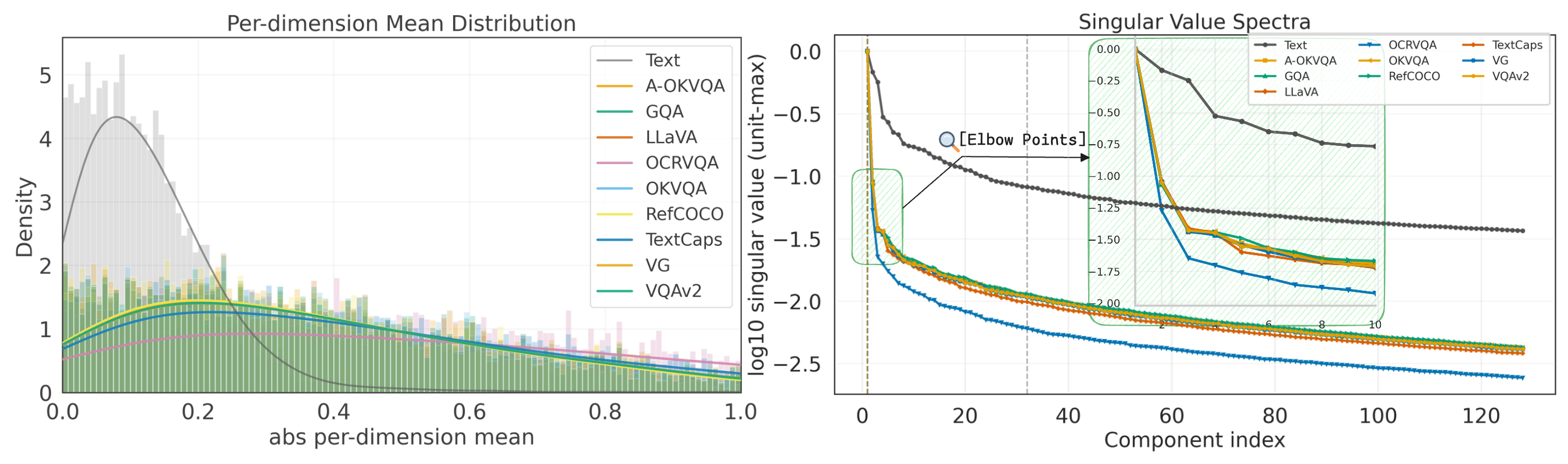}
    \caption{
    \textbf{Visual Diagnosis of Representation Anisotropy in Pre-trained MLLM Features.}
    \textbf{(Left)} The per-dimension mean distribution reveals a fundamental disparity: unlike text features \textcolor{grey}{(grey)} which are well-centered around zero, all visual instruction datasets exhibit a significant non-zero mean. This provides direct evidence that visual embeddings occupy a biased, narrow cone, which is the geometric origin of \textit{Global Semantic Drift}.
    \textbf{(Right)} The singular value scree plot confirms this diagnosis. The sharp "elbow point" indicates that the feature variance is confined to a few dominant dimensions—a classic symptom of representation degeneration. Detailed statistics and qualitative case studies are provided in Appendix~\ref{sec:appendix_anisotropy_stats} and Fig.~\ref{fig:case_study}.
}
    \label{fig:miu}
\end{figure*}
The advent of MLLMs marks a significant milestone in artificial intelligence, endowing models with remarkable capabilities to process and reason over interleaved visual and textual information \citep{liu2024improvedbaselinesvisualinstruction, zhu2023minigpt4enhancingvisionlanguageunderstanding, dai2023instructblip,aniri2026fabrimaeitrustmyself,aniri2026opdvvisualonpolicyselfdistillation}. The standard paradigm for building these models is a two-stage process: (1) large-scale pre-training on web-scraped image-text pairs to establish foundational cross-modal alignment, followed by (2) visual instruction tuning to hone the model's ability to follow complex, task-oriented instructions.
While instruction tuning is critical for unlocking state-of-the-art performance, its effectiveness is increasingly challenged by the data it relies on. Publicly curated datasets are expanding exponentially but are often saturated with low-quality and redundant examples \citep{chen2024visionlanguagemodelstrongfilter,wei2023instructiongpt4200instructionparadigmfinetuning}. Naively training on this noisy data inflates computational costs and, paradoxically, can degrade final model performance. This predicament has created an urgent need for efficient visual instruction selection methods that can identify a compact, high-quality subset of data.

Existing selection strategies predominantly fall into two categories. The first, \textbf{Proxy-Based Selection}, employs external models such as specialized scorers \citep{chen2024visionlanguagemodelstrongfilter} or auxiliary MLLMs \citep{lee2024conceptskilltransferabilitybaseddataselection} to estimate the importance of each data point. This approach, however, introduces substantial computational overhead from the proxy model's training and inference demands and is susceptible to selection bias when its objectives or architecture misalign with the target model. The second category, \textbf{Training-Based Selection}, leverages signals from the target model's own training dynamics, using metrics based on training loss \citep{liu2024morehighvaluedataselection} or influence functions \citep{wu2025iconsinfluenceconsensusvisionlanguage}. While this avoids proxy misalignment, these methods are often prohibitively expensive, requiring iterative gradient computations that can be as costly as training on the full dataset itself.

A critical limitation unites both paradigms: their substantial computational costs often negate the intended efficiency gains and can even lead to performance degradation under practical resource constraints. This reveals a fundamental gap in the current landscape—the lack of a selection method that is both effective at identifying high-quality data and genuinely efficient to deploy.
In this work, we argue that the key to bridging this gap lies not in designing more complex selection architectures, but in diagnosing and correcting a fundamental flaw within the MLLM's own visual representations. Our investigation begins with an analysis of the pre-trained visual feature distribution, where we uncover a critical, yet previously overlooked, property: \textit{representation anisotropy}. 

As illustrated in Figure~\ref{fig:miu}, we first observe that unlike text features, the mean embedding vector of visual features across diverse datasets is consistently non-zero. This signifies that the representations are confined to a narrow cone in high-dimensional space, a geometric bias that systematically distorts common similarity metrics \citep{levi2024double}. This distortion gives rise to a phenomenon we term \textit{Global Semantic Drift}, where global background context from different visual domains misleadingly inflates cosine similarity scores. A deeper analysis of the feature distribution's singular values confirms this profound anisotropy, a classic indicator of representation degeneration \citep{gao2019representationdegenerationproblemtrainingarxiv}. We posit that this anisotropy masks the intrinsic capability of the pre-trained features to discern data redundancy, as in such a space, geometric distance is no longer a faithful proxy for semantic similarity. We argue that the failure of prior methods to account for this anisotropy is a primary source of their excessive data selection overhead. Therefore, to effectively measure data redundancy, it is essential to first re-center the feature distribution to mitigate this anisotropic bias and reveal the underlying semantic structure.

Building on this insight, we introduce \textbf{PRISM}, a novel, training-free framework that redefines multimodal data selection. PRISM bypasses the complexities of existing proxy-based and training-based methods by establishing a third paradigm we term \textbf{Intrinsic Selection}. The philosophy of this paradigm is to unlock the latent potential of an MLLM's pre-trained representations by correcting, rather than circumventing, the flawed geometry of its feature space. 
At its core, PRISM counteracts representation anisotropy by performing an \textit{implicit re-centering} of the visual feature distribution. It effectively nullifies the global, non-semantic shift in the raw embeddings. This re-centering corrects a well-behaved isotropic geometry to the feature space. Crucially, it is this restoration that allows the \textbf{intrinsic features} themselves to become a direct and reliable signal for identifying visual semantic redundancy. With the geometric distortions removed, there is no need for external models or expensive gradient computations, enabling highly effective visual instruction selection with negligible computational cost.
We validate PRISM through extensive experiments on a diverse set of multimodal benchmarks, evaluating its efficacy against state-of-the-art selection methods. Our results demonstrate that an MLLM fine-tuned on a small, PRISM-selected dataset (\texttt{PRISM-Instruct-250K}) outperforms models trained on the full dataset while reducing training time by 70\%. On VisionFlan, PRISM selects only 57K samples from 186K examples and still exceeds full-data training. Furthermore, we conduct analyses on \textit{Cross-Model Generalization} and \textit{Knowledge Retention}, demonstrating that PRISM generalizes effectively across different MLLM architectures and better preserves the model's linguistic capabilities compared to full-dataset training.

\textbf{Our key contributions are as follows:}
\begin{itemize}
    \item We identify \textit{representation anisotropy} in the visual features of MLLMs as the root cause of inefficient data selection and diagnose its primary symptom, a \textit{Global Semantic Drift}, which distorts similarity metrics.
    \item We propose \textbf{PRISM}, a training-free framework that establishes a new \textbf{Intrinsic Selection} paradigm. By performing an \textit{implicit re-centering} of visual features, PRISM corrects an isotropic geometry, unlocking the MLLM's own features for redundancy detection.
    \item Our experiments show PRISM sets a new state-of-the-art, reducing training time by 70\% while yielding performance superior to models fine-tuned on the complete dataset.
\end{itemize}

\section{Visual Instruction Selection} 
\label{principle}
While numerous methods exist for visual instruction selection, a rigorous framework for evaluating their practical utility is often overlooked. Here we establish two fundamental principles that any effective selection strategy must satisfy to be considered viable in real-world applications. These principles serve as the foundation for a new quantitative metric to holistically assess selection efficacy.

\begin{principlebox}{Principle 1: Performance Fidelity}
A selected data subset must yield model performance equal to or greater than that achieved with the full dataset.
\end{principlebox}

\begin{principlebox}{Principle 2: Net Efficiency Gain}
The total time consumed by data selection and subsequent model tuning must be strictly less than the time required for tuning on the full dataset.
\end{principlebox}

\begin{figure}[t!]
    \centering
    \includegraphics[width=1\linewidth, trim={0.08cm 0.05cm 0.08cm 0.05cm}, clip]{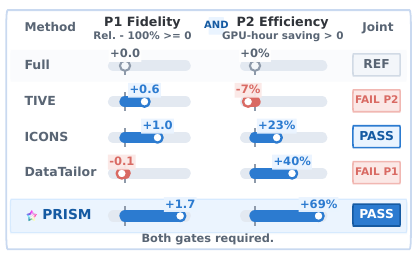}
    \caption{
    \textbf{Two-principle feasibility check.} A practical selector must satisfy both performance fidelity and net efficiency gain. PRISM is the only method that preserves full-data performance while substantially reducing end-to-end GPU hours.
    }
    \label{fig:osc}
\end{figure}
To unify these principles into a single, quantifiable measure, we introduce the \textbf{Overall Selection Cost (OSC)} score. This metric captures the total overhead of a selection method in terms of both performance and computation.
Formally, let $\mathcal{D}_\text{full}$ be the full instruction tuning dataset and $\mathcal{D}_\text{sub}$ be the selected subset. Let $P(\mathcal{D})$ denote the final performance score on the downstream benchmark achieved by a model fine-tuned on a dataset $\mathcal{D}$. The fine-tuning time on the full dataset is $T_\text{tune}(\mathcal{D}_\text{full})$, while the total time for a selection pipeline is the sum of the selection time $T_\text{select}$ and the subset fine-tuning time $T_\text{tune}(\mathcal{D}_\text{sub})$.
The Overall Selection Cost (OSC) score, $\mathcal{C}$, is defined as:
\begin{equation}
\label{eq:osc}
\begin{aligned}
\mathcal{C}
&=
\underbrace{
  \left(
    \frac{P(\mathcal{D}_{\mathrm{full}})}
         {P(\mathcal{D}_{\mathrm{sub}})}
  \right)
}_{\text{Performance Ratio}}
\\
&\quad \times
\underbrace{
  \left(
    \frac{
      T_{\mathrm{select}}
      + T_{\mathrm{tune}}(\mathcal{D}_{\mathrm{sub}})
    }{
      T_{\mathrm{tune}}(\mathcal{D}_{\mathrm{full}})
    }
  \right)
}_{\text{Time Ratio}} .
\end{aligned}
\end{equation}

A selection method is considered practically viable if and only if its OSC score $\mathcal{C} < 1$. This single criterion guarantees that any computational savings are not achieved at the expense of final model performance, thus certifying a true net gain in efficiency as shown in Fig. \ref{fig:osc} and Fig. \ref{fig:osc1}.

\begin{figure*}[ht]
  \centering
  \includegraphics[width=\linewidth]{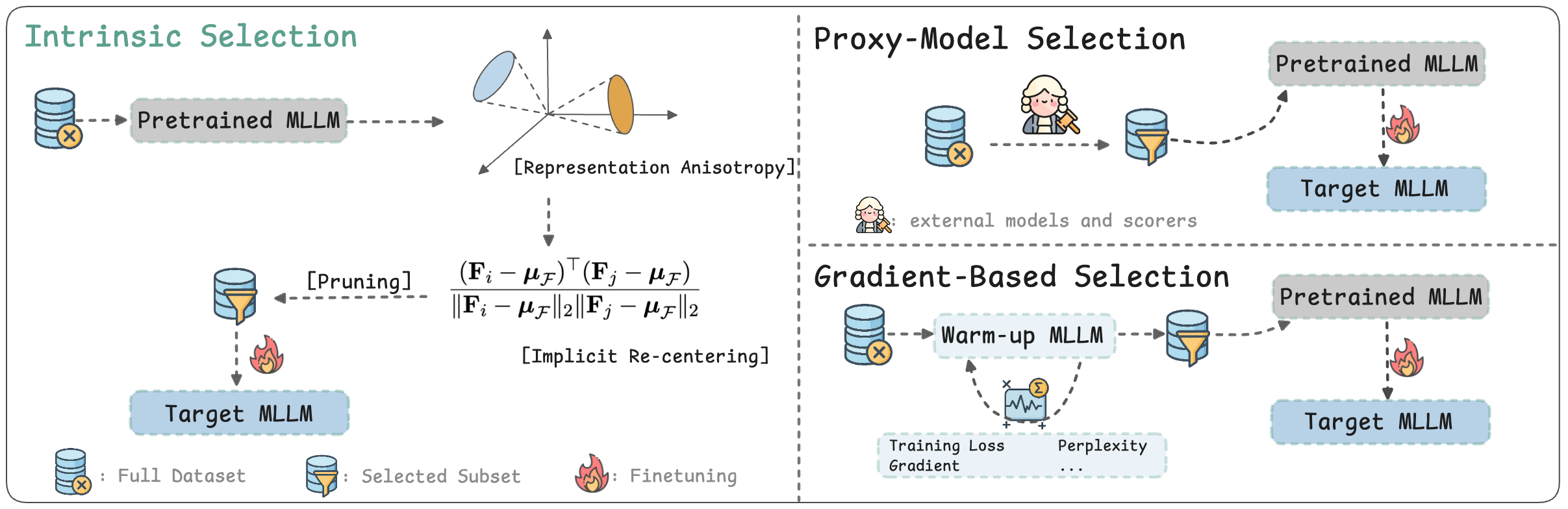}
\caption{
    \textbf{A Comparison of Data Selection Paradigms.} Existing methods, such as Proxy-Based and Training-Based Selection, introduce significant computational overhead by relying on external scorers or expensive, iterative training loops to estimate data importance. In contrast, our proposed \textbf{Intrinsic Selection} paradigm operates on a more fundamental principle: we first diagnose a geometric flaw—\textit{representation anisotropy}—in the MLLM's native visual features. We then apply an \textit{implicit re-centering} to correct this bias. This isotropic restoration unlocks the intrinsic semantic structure of the data, enabling highly effective and training-free pruning in a single pass.
}
  \label{fig:Visual_Instruction_Selection}
\end{figure*}

As illustrated in Fig.~\ref{fig:Visual_Instruction_Selection}, existing visual instruction selection methods can be broadly categorized into two main paradigms. The first is \textbf{Proxy-Based Selection}, where the target MLLM remains a black box, and data quality is evaluated using an external scoring function. This function can be a purpose-built scoring model \citep{chen2024visionlanguagemodelstrongfilter}, a human reward model, or a powerful LLM-based scorer \citep{wei2023instructiongpt4200instructionparadigmfinetuning}. Some approaches also train a smaller proxy MLLM to guide the data selection for a larger target model \citep{lee2024conceptskilltransferabilitybaseddataselection}.
The second category is \textbf{Training-Based Selection}. In this approach, the target MLLM is actively used to assess data value by computing metrics derived from its training dynamics, such as training loss, perplexity, or gradient information on a given data partition.
Yet both paradigms exhibit structural drawbacks: they risk \textbf{performance degradation}, incur \textbf{high computational costs}, and introduce \textbf{Proxy-Based bias}. A fuller discussion appears in Appendix~\ref{sec:limitations_and_motivation}.
We contend that these drawbacks share a single, underappreciated root cause: the overlooked \emph{representation anisotropy} inherent in MLLM visual feature distributions. To address this geometric pathology, we introduce \textbf{PRISM}—a self-\textbf{PR}uning \textbf{I}ntrinsic \textbf{S}election \textbf{M}ethod—that enables training-free multimodal data selection by operating directly on the model’s intrinsic, re-centered features.
\section{PRISM}
\prismtightspacing
\subsection{Representation Anisotropy}
To reveal the potential of pre-trained intrinsic features, we first investigate the feature distribution and uncover a critical, yet previously overlooked property in previous data selection methods: representation anisotropy.
Representation anisotropy is a geometric pathology observed in many high-dimensional embedding spaces \citep{levi2024double,zhang-etal-2020-unsupervised}.  
In an \emph{isotropic} embedding, vectors are uniformly distributed around the origin and exhibit equal variance along every dimension; mathematically, the covariance matrix of an isotropic set is proportional to the identity.  Such a space fully exploits its dimensionality and provides maximal directional diversity for encoding semantic concepts. 
As illustrated in Figure~\ref{fig:miu}, we begin with a deep analysis of the feature geometry from a pre-trained MLLM, revealing a critical flaw in its visual representations. We find a stark disparity between modalities: while textual features are well-centered around a zero mean, the feature distributions for all visual datasets exhibit a significant, non-zero mean. This confirms that visual embeddings are confined to a narrow cone in high-dimensional space—a geometric bias that systematically distorts similarity metrics and gives rise to a phenomenon we term \textit{Global Semantic Drift}, where shared background context is misinterpreted as semantic content.
A singular value analysis of these distributions further confirms this profound anisotropy. The sharp "elbow point" observed in Figure~\ref{fig:miu} (right) indicates that feature variance is concentrated in only a few dominant dimensions, a classic symptom of representation degeneration that corrupts semantic distance measures \citep{levi2024double}. We posit that this unaddressed anisotropy is the primary reason for the inefficiency of prior art.

\subsection{Theoretical Analysis: Anisotropy as a Source of Computational Overhead}
\label{sec:theory}
To formalize this, let $\phi: \mathcal{D} \to \mathbb{R}^d$ be an embedding function mapping data points to representation vectors. The observed anisotropy implies a non-zero mean vector $\boldsymbol{\mu} = \mathbb{E}[\phi(d)]$ with significant magnitude. We can thus decompose any representation $\mathbf{x}_i = \phi(d_i)$ into a shared global component and a sample-specific residual:
\begin{equation}
    \mathbf{x}_i = \boldsymbol{\mu} + \boldsymbol{\delta}_i
\end{equation}
where $\boldsymbol{\delta}_i$ encodes the unique semantics of sample $d_i$. The defining characteristic of anisotropy is that the shared component dominates, i.e., $\|\boldsymbol{\mu}\|_2 \gg \mathbb{E}[\|\boldsymbol{\delta}_i\|_2]$, which is empirically substantiated in Appendix~\ref{sec:appendix_anisotropy_stats}. This geometric property has a catastrophic consequence for standard metrics.

\newtheorem{theorem}{Theorem}
\begin{theorem}
\label{thm:contamination}
In an anisotropic representation space where $\|\boldsymbol{\mu}\|_2 \gg \mathbb{E}[\|\boldsymbol{\delta}_i\|_2]$, the cosine similarity between any two randomly sampled vectors $\mathbf{x}_i$ and $\mathbf{x}_j$ is dominated by the global drift $\boldsymbol{\mu}$, masking their semantic dissimilarity encoded in $\boldsymbol{\delta}_i, \boldsymbol{\delta}_j$. 
\end{theorem}
\begin{proof}
The cosine similarity is given by:
\begin{equation}
\cos(\mathbf{x}_i, \mathbf{x}_j) = \frac{(\boldsymbol{\mu} + \boldsymbol{\delta}_i)^\top (\boldsymbol{\mu} + \boldsymbol{\delta}_j)}{\|\boldsymbol{\mu} + \boldsymbol{\delta}_i\|_2 \|\boldsymbol{\mu} + \boldsymbol{\delta}_j\|_2}
\end{equation}
The numerator expands to $\|\boldsymbol{\mu}\|_2^2 + \boldsymbol{\mu}^\top(\boldsymbol{\delta}_i + \boldsymbol{\delta}_j) + \boldsymbol{\delta}_i^\top\boldsymbol{\delta}_j$. The denominator can be approximated using a first-order Taylor expansion $\sqrt{a+b} \approx \sqrt{a} + \frac{b}{2\sqrt{a}}$ for small $b$. As $\|\boldsymbol{\mu}\|_2$ is the dominant term, we have $\|\boldsymbol{\mu} + \boldsymbol{\delta}_i\|_2 \approx \|\boldsymbol{\mu}\|_2 + \frac{\boldsymbol{\mu}^\top\boldsymbol{\delta}_i}{\|\boldsymbol{\mu}\|_2}$. Substituting these into the main equation and retaining only the dominant terms yields:
\begin{align}
\cos(\mathbf{x}_i, \mathbf{x}_j)
&\approx 
\frac{\|\boldsymbol{\mu}\|_2^2}{
\left(\|\boldsymbol{\mu}\|_2 + \frac{\boldsymbol{\mu}^\top\boldsymbol{\delta}_i}{\|\boldsymbol{\mu}\|_2}\right)
\left(\|\boldsymbol{\mu}\|_2 + \frac{\boldsymbol{\mu}^\top\boldsymbol{\delta}_j}{\|\boldsymbol{\mu}\|_2}\right)} \nonumber\\
&\approx 
1 - \frac{1}{2}\left\|
\frac{\boldsymbol{\delta}_i'}{\|\boldsymbol{\mu}\|_2}
- 
\frac{\boldsymbol{\delta}_j'}{\|\boldsymbol{\mu}\|_2}
\right\|_2^2 
\approx 1
\end{align}
where $\boldsymbol{\delta}'$ is the component of $\boldsymbol{\delta}$ orthogonal to $\boldsymbol{\mu}$.
The true semantic similarity, proportional to $\boldsymbol{\delta}_i^\top\boldsymbol{\delta}_j$, is relegated to lower-order terms, rendering the metric ineffective for distinguishing unique samples. A full derivation is provided in Appendix \ref{sec:appendix_theory}.
\end{proof}

\newtheorem{corollary}[theorem]{Corollary}
\begin{corollary}
\label{cor:failure}
Due to Theorem \ref{thm:contamination}, computationally efficient data selection algorithms predicated on geometric dissimilarity in the raw embedding space (e.g., Farthest-Point Sampling, distance-based clustering) fail to select semantically diverse subsets.
\end{corollary}

\noindent This failure of inexpensive geometric proxies forces existing methods to adopt computationally intensive strategies to approximate the true, underlying data value, effectively paying a high computational ``tax'' for ignoring the geometric flaw of the space.

\subsection{Redundancy Identification via Implicit Re-centering of Intrinsic Features}

\noindent To solve the Representation Anisotropy and alleviate Global Semantic Drift, PRISM employs a implicit re-centering operation to restore a well-behaved isotropic geometry to the feature space. Crucially, it is this restoration that allows the intrinsic features themselves to become a direct and reliable signal for identifying semantic redundancy.

Let our dataset be $\mathcal{D} = \{d_1, \dots, d_N\}$. For each data point $d_i$, we first extract a high-level feature representation $\mathbf{F}_i \in \mathbb{R}^d$ from an intermediate layer of pretrained MLLM. The set of these vectors, $\{\mathbf{F}_i\}_{i=1}^N$, forms the anisotropic space diagnosed in Section \ref{sec:theory}, characterized by the non-zero global mean vector $\boldsymbol{\mu}_{\mathcal{F}} = \frac{1}{N}\sum_{i=1}^N \mathbf{F}_i$.
To measure similarity in a geometrically corrected space where the influence of this global drift is nullified, we keep the normalization-and-inner-product form of a Pearson-style correlation coefficient while adapting its re-centering term to the dataset-level mean.
Specifically, we define the centered vector as $\mathbf{F}'_i = \mathbf{F}_i - \boldsymbol{\mu}_{\mathcal{F}}$, where $\boldsymbol{\mu}_{\mathcal{F}}$ captures the shared global shift across the corpus.
\begin{equation}
\label{eq:pearson_as_cosine}
\rho(\mathbf{F}_i, \mathbf{F}_j) = \frac{(\mathbf{F}_i - \boldsymbol{\mu}_{\mathcal{F}})^\top (\mathbf{F}_j - \boldsymbol{\mu}_{\mathcal{F}})}{\|\mathbf{F}_i - \boldsymbol{\mu}_{\mathcal{F}}\|_2 \|\mathbf{F}_j - \boldsymbol{\mu}_{\mathcal{F}}\|_2} 
\end{equation}
This operation measures the alignment of two vectors' internal variation patterns after removing the global, non-semantic shift that characterizes anisotropy. By operating on these centered vectors $\mathbf{F}'_i$, we effectively evaluate the relationship between the unique semantic components, $\boldsymbol{\delta}_i$, that our theory identified.

With a robust similarity metric established, we quantify the redundancy of each sample. A sample is considered redundant if its unique semantic information is highly correlated with many other samples in the dataset. We define the \textit{Redundancy Score} $\mathcal{R}(d_i)$ for a sample $d_i$ as its expected correlation with the rest of the corpus:
\begin{equation}
    \label{eq:redundancy_score}
    \mathcal{R}(d_i) = \frac{1}{N-1} \sum_{j=1, j \neq i}^{N} \rho(\mathbf{F}_i, \mathbf{F}_j)
\end{equation}
 $\mathcal{R}(d_i)$ is a corpus-wide redundancy estimator: after re-centering, it measures the weighted degree of $d_i$ in the complete semantic-similarity graph induced by $\rho(\cdot,\cdot)$. The score is computed by an exact aggregate implementation in $\mathcal{O}(Nd)$ time without materializing any $N \times N$ matrix; details in Appendix~\ref{sec:runtime_details}. Given a selection budget defined by a percentile threshold $\tau$, we select all samples whose redundancy scores fall below the $\tau$-th percentile of the score distribution:
\begin{equation}
    \mathcal{D}_{\text{select}} = \left\{ d_i \in \mathcal{D} \mid \mathcal{R}(d_i) \leq Q_{\tau}\left(\{\mathcal{R}(d_k)\}_{k=1}^N\right) \right\},
\end{equation}
where $Q_{\tau}(\cdot)$ is the function that computes the $\tau$-th percentile. This principled approach ensures that the selected subset is enriched with semantically diverse and informative samples, having explicitly corrected for the geometric distortions of the original embedding space.

\begin{algorithm}[t]
    \caption{PRISM}
    \label{algo:prism}
    \begin{algorithmic}[1]
        \State \textbf{Input:} Dataset $\mathcal{D} = \{d_1, \dots, d_N\}$, MLLM components ($\text{VE}, \text{Proj}, \text{LLM}^{(l)}$), percentile $\tau$
        \State \textbf{Output:} Selected subset $\mathcal{D}_{\text{select}}$
        \State \textcolor{stepblue}{\textbf{Anisotropic Feature Extraction:}}
        \ForAll{$d_i = (I_i, T_i) \in \mathcal{D}$}
            \State $v_i \gets \text{VE}(I_i)$ \hfill 
            \State $z_i \gets \text{Proj}(v_i)$ \hfill 
            \State $\mathbf{F}_i \gets \frac{1}{T} \sum_{t=1}^{T} \text{LLM}^{(l)}(z_i)_t$ \hfill \textcolor{commentgreen}{/* Obtain layer-$l$ representation */}
        \EndFor
        \State \textcolor{stepblue}{\textbf{Redundancy Identification via Implicit Re-centering:}}
        \State $\boldsymbol{\mu}_{\mathcal{F}} \gets \frac{1}{N}\sum_{k=1}^{N}\mathbf{F}_k$
        \ForAll{$i \in \{1, \dots, N\}$}
            \State $\mathcal{R}_i \gets \frac{1}{N-1} \sum_{j \neq i} \frac{(\mathbf{F}_i - \boldsymbol{\mu}_{\mathcal{F}})^\top (\mathbf{F}_j - \boldsymbol{\mu}_{\mathcal{F}})}{\|\mathbf{F}_i - \boldsymbol{\mu}_{\mathcal{F}}\|_2 \|\mathbf{F}_j - \boldsymbol{\mu}_{\mathcal{F}}\|_2}$ \hfill \textcolor{commentgreen}{/* Average correlation is implemented by an exact aggregate form in $\mathcal{O}(Nd)$ */}
        \EndFor
        \State \textcolor{stepblue}{\textbf{Diversity-Aware Pruning:}}
        \State $q_{\tau} \gets \text{Percentile}(\{\mathcal{R}_k\}_{k=1}^N, \tau)$ \hfill \textcolor{commentgreen}{/* Find score threshold at $\tau$-th percentile */}
        \State $\mathcal{D}_{\text{select}} \gets \{ d_i \in \mathcal{D} \mid \mathcal{R}_i \leq q_{\tau} \}$ \hfill \textcolor{commentgreen}{/* Filter samples with low redundancy scores */}
        \State \textbf{return} $\mathcal{D}_{\text{select}}$
    \end{algorithmic}
\end{algorithm}

\section{Experiments}
\begin{table*}[t!]
\centering
\resizebox{\textwidth}{!}{ % 调节表格宽度为文本宽度，高度按比例缩放
\begin{tabular}{l*{12}{c}}
\toprule
\textbf{Method} & \textbf{SQA} & \textbf{SQA-I} & \textbf{VizWiz}& \textbf{POPE-P} & \textbf{POPE-R} & \textbf{POPE-A} & \textbf{MM-Vet} & \textbf{MMBench} & \textbf{MME-P} & \textbf{MME-C} & \textbf{MMMU} & \textbf{\cellcolor{gray!10}Rel. (\%)} \\
\midrule
Full-Finetune & 69.4 & 66.8& 50.0 & 86.1& 87.3&84.2 &31.1 &64.3 &1510.7 & 311.9& 35.4& 100\% \cellcolor{gray!10} \\
\cmidrule(lr){1-13} % 在 Full-Finetune 下面加一条分割线
Random &65.5 &64.5 &48.1  &85.1 &84.6 &83.6 &30.2 &55.5 &1492.0 &233.5 &30.5 &93.2\% \cellcolor{gray!10} \\
Length &66.8 &66.7 & 47.0 &85.4 &85.5 &84.1 &31.5 &57.0 & 1422.1& 306.0&33.1 & 96.6\%\cellcolor{gray!10} \\
EL2N &70.2 &70.6 & 44.4& 85.6 & 85.6 &85.6 &- &61.6 &1356.5 &294.7 & -&97.2\% \cellcolor{gray!10} \\
Perplexity &70.5 & 67.9& -& 83.3  & 83.3 &83.3  &- & 62.3&1393.3 & 260.7& -&95.8\% \cellcolor{gray!10} \\
GraNd &71.4 &68.4& 37.8& 82.5 & 82.5 &82.5 &- & 62.9&  1400.5&287.1& -& 94.6\%\cellcolor{gray!10} \\
TIVE &72.2 &70.6 & -& 85.6 & 85.6 &85.6 &- &63.2 &1433.0 &322.1 & -&100.6\% \cellcolor{gray!10} \\
InstructionGPT-4 & -& -& -& - & -&- &- &31.4 &463.3 &- & -&39.75\% \cellcolor{gray!10} \\
Self-Filter &- & 61.4& 53.2& 83.8 & 83.8&83.8 &- &61.4 & 1306.2& -& -&96.1\% \cellcolor{gray!10} \\
COINCIDE & -& 69.2& 46.8& 86.1 & 86.1&86.1 &- &63.1 &1495.6 &- & -&99.3\% \cellcolor{gray!10} \\
ICONS &- &70.8 & -& 87.5 & 87.5&87.5 &- &63.1 &1485.7& - & -& 101.0\%\cellcolor{gray!10} \\
DataTailor  & 71.0& -& 49.5& 85.3 & 85.3&85.3 &- &- & 1476.1& 319.2& -&99.9\% \cellcolor{gray!10} \\
\includegraphics[height=1em]{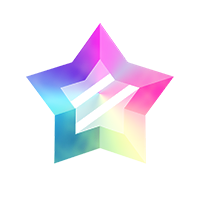}\cellcolor{cyan!10}PRISM (Ours) & \textbf{71.3}\cellcolor{cyan!10} &\textbf{69.1} \cellcolor{cyan!10}  & \textbf{50.1}\cellcolor{cyan!10} &\textbf{87.7} \cellcolor{cyan!10} &\textbf{88.7} \cellcolor{cyan!10} &\textbf{85.5} \cellcolor{cyan!10} & \textbf{32.0}\cellcolor{cyan!10} &\textbf{65.2} \cellcolor{cyan!10} & 1470.0\cellcolor{cyan!10} &\textbf{330.0} \cellcolor{cyan!10} & 34.7\cellcolor{cyan!10} & \textbf{101.7\%}\cellcolor{cyan!10} \\
\bottomrule
\end{tabular}
}
\caption{Evaluation of PRISM against full fine-tuning and existing data selection approaches across multiple multimodal understanding benchmarks. PRISM achieves better performance, surpassing full fine-tuning while significantly reducing computational costs. Metrics in \textbf{bold} indicate improvements over the full fine-tuning baseline. For POPE, we report the average score across three subsets for certain baselines due to the unavailability.}
\label{tab:main}
\end{table*}

\begin{figure*}
    \centering
    \includegraphics[width=1\linewidth]{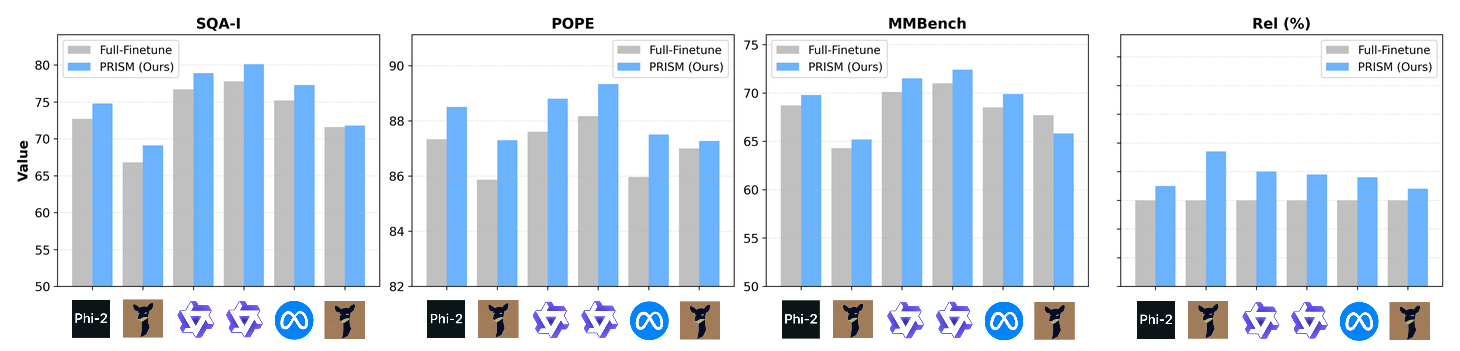}
  \caption{\textbf{Generalization across diverse MLLM families and scales.} Comparison of PRISM (blue) and full fine-tuning (grey) across SQA-I, POPE, MMBench, and Rel. (\%) (detailed numerical results provided in Table~\ref{tab:arch} ). Results for \textbf{Phi-2}, \textbf{Vicuna}, \textbf{Llama-3}, and \textbf{Qwen2.5} confirm that PRISM's performance gains are architecture-agnostic and transfer robustly across varied backbones and vision encoders.}
    \label{fig:archnew}
\end{figure*}

\subsection{Experiment Setup}
\textbf{Dataset \& Model:} We evaluate PRISM on the visual instruction tuning dataset LLaVA-665K \citep{liu2024improvedbaselinesvisualinstruction}, using LLaVA-1.5-7B \citep{liu2024improvedbaselinesvisualinstruction} as our primary base model. To test robustness beyond the LLaVA-style mixture, we additionally evaluate PRISM on VisionFlan-186K \citep{xu-etal-2024-vision}, a distinct instruction pool containing 191 vision-language tasks and 186K samples. All experiments are conducted for one epoch following the official fine-tuning hyperparameters. To ensure a fair comparison, we maintain a consistent training environment across all evaluations.

\noindent\textbf{Baselines:} We compare PRISM against a comprehensive set of baselines, including Random Selection, Instruction Length, Perplexity \citep{liu2024morehighvaluedataselection}, GraNd \citep{paul2023deeplearningdatadiet}, EL2N \citep{paul2023deeplearningdatadiet}, InstructionGPT-4 \citep{wei2023instructiongpt4200instructionparadigmfinetuning}, SELF-FILTER \citep{chen2024visionlanguagemodelstrongfilter}, TIVE \citep{liu2024morehighvaluedataselection}, COINCIDE \citep{lee2024conceptskilltransferabilitybaseddataselection}, DataTailor \citep{yu2024mastering}, and ICONS \citep{wu2025iconsinfluenceconsensusvisionlanguage}. To ensure fair comparisons, we adopt the experimental settings and incorporate results from ICONS \citep{wu2025iconsinfluenceconsensusvisionlanguage} and TIVE \citep{liu2024morehighvaluedataselection}.

\noindent\textbf{Benchmarks.}
Following the rigorous framework of LLaVA-1.5 \citep{liu2024improvedbaselinesvisualinstruction}, we evaluate PRISM across a comprehensive suite of seven diverse benchmarks. This suite is designed to probe a wide array of MLLM capabilities, including fine-grained perception, visual reasoning, hallucination detection, and complex conversational skills. Our evaluation includes MMBench \citep{liu2024mmbenchmultimodalmodelallaround}, ScienceQA \citep{lu2022learn}, MME \citep{yin2023survey}, POPE \citep{li2023evaluatingobjecthallucinationlarge}, VizWiz \citep{gurari2018vizwizgrandchallengeanswering}, MM-Vet \citep{yu2024mmvetevaluatinglargemultimodal}, and MMMU \citep{yue2024mmmumassivemultidisciplinemultimodal}. Detailed descriptions of each benchmark are provided in Appendix~\ref{sec:benchmark_details}.

\subsection{Main Results}
\begin{figure}
    \centering
    \includegraphics[width=1\linewidth, trim={3.2cm 11cm 18cm 2.5cm}, clip]{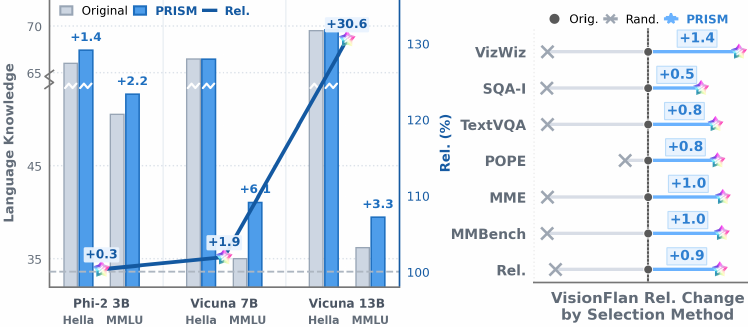}
    \caption{PRISM preserves language knowledge and robustly generalizes to VisionFlan-186K  Instruction Pool.}
    \label{fig:flan_llm_summary}
\end{figure}

We present a comprehensive evaluation to validate PRISM's effectiveness and robustness. Our primary results in Table~\ref{tab:main} demonstrate that PRISM outperforms both full fine-tuning and state-of-the-art selection methods on the LLaVA-1.5-7B model. We then showcase PRISM's broad applicability by demonstrating its strong performance across different MLLM architectures (Table~\ref{tab:arch}), in uni-modal, text-only settings (Fig.~\ref{fig:flan_llm_summary} and Table~\ref{tab:text}), and on the distinct VisionFlan instruction pool (Fig.~\ref{fig:flan_llm_summary} and Appendix~\ref{sec:visionflan_robustness}).
We highlight the following findings:

\exppoint{172}{State-of-the-Art Performance with Less Data.} As shown in Table~\ref{tab:main}, PRISM achieves superior performance across various benchmarks, outperforming full-dataset fine-tuning by a relative \textbf{1.7\%}. The gains are particularly notable on complex reasoning tasks like MMBench (65.2 vs. 64.3) and MM-Vet (32.0 vs. 31.1). This demonstrates that PRISM preferentially selects high-value, instruction-rich data more effectively than existing methods.

\exppoint{173}{Cross-Model Generalization and Scalability.} PRISM's selection mechanism generalizes robustly across diverse MLLM architectures by correcting representation anisotropy. By identifying semantically dense samples, we establish the approach's breadth across five model--encoder combinations. Results demonstrate that effectiveness holds for both small and large language models—including Vicuna, Phi-2, Qwen2.5, and Llama-3—and for distinct vision encoders like CLIP and SigLIP (details in Appendix~\ref{tab:model_arch}). These findings, illustrated in Fig.\ref{fig:archnew}, confirm that PRISM’s gains are not specific to the LLaVA architecture but transfer effectively across diverse multimodal families.

\exppoint{174}{Balanced Efficiency and Performance.}
As illustrated in Figure~\ref{fig:efficiency_pareto}, PRISM occupies the most favorable efficiency--performance region: it improves over full fine-tuning while substantially reducing end-to-end GPU hours. This advantage comes from its training-free design and aggregate correlation computation. A detailed breakdown of the wall-clock runtime can be found in Appendix~\ref{sec:runtime_details}.

\exppoint{175}{Robustness to Instruction-Pool Shift.}
Beyond LLaVA-665K, PRISM also generalizes to VisionFlan-186K, a more task-diverse instruction pool with 191 distinct vision-language tasks. With the same 30\% selection budget, PRISM selects 57K samples and reaches 100.9\% relative performance over full-data training, substantially outperforming random selection at the same budget (94.1\%). The comparison is summarized in Fig.~\ref{fig:flan_llm_summary}, with full results in Table~\ref{tab:visionflan}. This confirms that PRISM's geometric selection principle is not tied to a single LLaVA-style data mixture.

\subsection{In-Depth Model Behavior Analysis}

\exppoint{170}{Effective Hallucination Mitigation.} PRISM sets a new state-of-the-art on all POPE subsets for hallucination detection, surpassing even specialized methods. This result suggests that our core mechanism—alleviating representation anisotropy via implicit re-centering —is highly effective at removing samples containing spurious semantic drift, which are a known cause of object hallucination.

\begin{table*}[t!]
\centering
\resizebox{\textwidth}{!}{ % 调节表格宽度为文本宽度，高度按比例缩放
\begin{tabular}{l*{12}{c}}
\toprule
\textbf{Method} & \textbf{SQA} & \textbf{SQA-I} & \textbf{VizWiz} & \textbf{POPE-P} & \textbf{POPE-R} & \textbf{POPE-A} & \textbf{MM-Vet} & \textbf{MMBench} & \textbf{MME-P} & \textbf{MME-C} & \textbf{MMMU} & \textbf{Rel. (\%)} \\
\midrule
Deep Layer & 71.2& 69.1& 51.6& 86.6& 88.0&84.2 &31.1 & 62.9&1477.0 &254.0 &34.5 & 97.2\%\\

Middle Layer &70.9 &69.1 &47.7 &86.5 &87.8 &84.2 &31.9 &65.0 &1517.1 &276.0 & 34.9&97.9\% \\

\raisebox{-0.2em}{\includegraphics[height=0.8em]{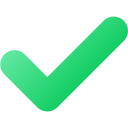}} Shallow Layer & \textbf{71.3}\cellcolor{cyan!10} &\textbf{69.1} \cellcolor{cyan!10}  & 50.1\cellcolor{cyan!10} &\textbf{87.7} \cellcolor{cyan!10} &\textbf{88.7} \cellcolor{cyan!10} &\textbf{85.5} \cellcolor{cyan!10} & \textbf{32.0}\cellcolor{cyan!10} &\textbf{65.2} \cellcolor{cyan!10} & 1470.0\cellcolor{cyan!10} &\textbf{330.0} \cellcolor{cyan!10} & 34.7\cellcolor{cyan!10} & \textbf{100.0\%}\cellcolor{cyan!10} \\
\cmidrule(lr){1-13} 

High Correlation &70.6&68.0 &48.1 &85.8 &87.6 &83.9 &30.7 &64.0 &1428.5 &275.3 &33.5 &96.3\%  \\

Moderate Correlation & 71.0&69.7 &48.3 &85.9 &86.7 &84.0 &30.0 &64.2 &1509.0 &286.0 & 34.1&97.3\% \\

\raisebox{-0.2em}{\includegraphics[height=0.8em]{figs/12483567.png}} Low Correlation & \textbf{71.3}\cellcolor{cyan!10} &69.1 \cellcolor{cyan!10}  & \textbf{50.1}\cellcolor{cyan!10} &\textbf{87.7} \cellcolor{cyan!10} &\textbf{88.7} \cellcolor{cyan!10} &\textbf{85.5} \cellcolor{cyan!10} & \textbf{32.0}\cellcolor{cyan!10} &\textbf{65.2} \cellcolor{cyan!10} & 1470.0\cellcolor{cyan!10} &\textbf{330.0} \cellcolor{cyan!10} & \textbf{34.7}\cellcolor{cyan!10} & \textbf{100.0\%}\cellcolor{cyan!10} \\
\cmidrule(lr){1-13} 

Last Token & 69.9&67.3 &49.4 &87.4 &88.3 &85.0 &31.6 &62.6 &1471.0 &272.0 &35.3 & 97.4\%\\

\raisebox{-0.2em}{\includegraphics[height=0.8em]{figs/12483567.png}} Avg Pooling & \textbf{71.3}\cellcolor{cyan!10} &\textbf{69.1} \cellcolor{cyan!10}  & \textbf{50.1}\cellcolor{cyan!10} &\textbf{87.7} \cellcolor{cyan!10} &\textbf{88.7} \cellcolor{cyan!10} &\textbf{85.5} \cellcolor{cyan!10} & \textbf{32.0}\cellcolor{cyan!10} &\textbf{65.2} \cellcolor{cyan!10} & 1470.0\cellcolor{cyan!10} &\textbf{330.0} \cellcolor{cyan!10} & 34.7\cellcolor{cyan!10} & \textbf{100.0\%}\cellcolor{cyan!10} \\
\bottomrule
\end{tabular}
}
\caption{
    \textbf{Ablation study validating the core design principles of PRISM.} The results confirm that performance is maximized by using Shallow Layer features (preserving a cleaner geometric structure), selecting for Low Correlation (maximizing informational diversity), and applying Average Pooling (ensuring a holistic representation). 
}
\label{tab:ablation}
\end{table*}

\exppoint{171}{Language Knowledge Retention.}
Visual instruction tuning is known to degrade a model's foundational language capabilities, a phenomenon often termed "knowledge forgetting" \citep{zhang2024wingslearningmultimodalllms}. We show that PRISM not only mitigates this effect but can even enhance text-only performance. We hypothesize this is because PRISM's Pearson-style correlation scoring effectively removes samples with spurious visual-textual associations. By filtering out this "noisy" data that often causes catastrophic forgetting, our method provides a cleaner training signal. As shown in Fig.~\ref{fig:flan_llm_summary} and detailed in Table~\ref{tab:text}, models trained on the PRISM subset consistently outperform their full-dataset-trained counterparts on language-centric benchmarks. This highlights a dual benefit of PRISM: achieving state-of-the-art multimodal performance without sacrificing language competency.

\begin{figure}[t!]
    \centering
    \includegraphics[width=1\linewidth]{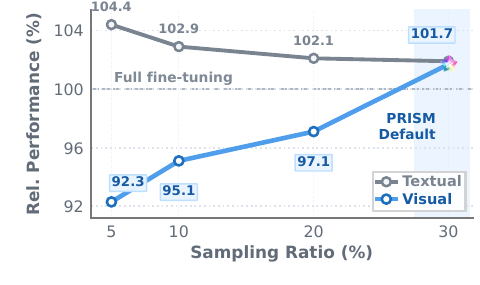}
    \caption{
        \textbf{Sampling-ratio trade-off in PRISM.} Visual performance (blue) increases with larger selected subsets and surpasses full fine-tuning at 30\%, while textual performance is best preserved under smaller budgets.
    }
    \label{fig:sampling_ratio_analysis}
\end{figure}

\begin{figure}
    \centering
    \includegraphics[width=1\linewidth]{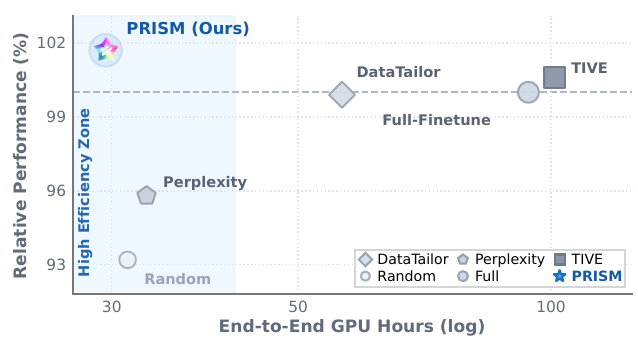}
    \caption{\textbf{Efficiency--performance Pareto comparison.} PRISM achieves the highest relative performance (101.7\%) while reducing total GPU time by about 70\% compared with full fine-tuning. TIVE attains comparable accuracy but exceeds the full fine-tuning cost, whereas cheaper heuristics such as Random and Perplexity sacrifice substantial performance.}
    \label{fig:efficiency_pareto}
\end{figure}

\exppoint{169}{Analysis of the Sampling Ratio.}
\label{sec:sampling_ratio}
We analyze PRISM under 4 sampling ratios to characterize the trade-off between data budget and dual-modality performance (Figure~\ref{fig:sampling_ratio_analysis}). The results show that PRISM remains robust across budgets, with larger subsets favoring visual performance and smaller subsets better preserving language knowledge.

\subsection{Ablation Study}
To validate the design choices of PRISM, we conduct ablation studies on three key components: the layer of visual features, the Pearson-style correlation scoring, and the aggregation strategy.

\exppoint{172}{Influence of LLM Layer Selection.}
We find that features from a \textbf{Shallow Layer} (Layer 1) are most effective for PRISM, significantly outperforming those from middle or deep layers (Table~\ref{tab:ablation}). This suggests that the fundamental geometric properties required for redundancy detection are most distinct in early layers, before complex abstractions introduce further representational artifacts.

\exppoint{173}{Impact of Correlation-based Selection.}
Validating our central hypothesis, we show that selecting the \textbf{Low-Correlation} data subset yields the highest performance. This confirms that our implicit re-centering strategy effectively identifies informational diversity and prunes redundant samples that provide diminishing returns for model training (see Appendix \ref{sec:limitations_and_motivation} for further analysis).

\exppoint{174}{Effect of Token Aggregation Strategy.}
We confirm that using an \textbf{Average} of all image tokens to form the global feature vector is superior to using only the \textbf{Last Image Token}. Averaging provides a more holistic and robust representation of an image's semantics, which is crucial for the stability of our geometric analysis, whereas relying on a single token can be susceptible to positional biases.

\section{Conclusion}
We trace MLLM selection inefficiency to representation anisotropy and propose \textbf{PRISM}, a training-free method that corrects semantic structure via implicit re-centering. \textbf{PRISM} reduces training time by 70\% while surpassing full-dataset performance.

\section*{Limitations}

While PRISM demonstrates significant promise, we acknowledge several limitations that offer avenues for future research.
Our current approach focuses specifically on identifying and pruning semantic redundancy based on feature correlation. It is not designed to detect other critical data issues, such as factual inaccuracies, ethical biases within the instruction data.
Future work could explore extending our geometry-aware principles to address these other facets of data quality or adapting the method to new modalities beyond vision and language.

\section*{Acknowledgements}

The authors gratefully acknowledge the scientific support and HPC resources provided by the Erlangen National High Performance Computing Center (NHR@FAU) of the Friedrich-Alexander-Universität Erlangen-Nürnberg (FAU) under the NHR project b270dd and b271dd. NHR funding is provided by federal and Bavarian state authorities. NHR@FAU hardware is partially funded by the German Research Foundation (DFG) – 440719683. 

We acknowledge the EuroHPC Joint Undertaking for awarding this project access to the EuroHPC supercomputer LEONARDO, hosted by CINECA (Italy) and the LEONARDO consortium through an EuroHPC AI and Data-Intensive Applications Access call EHPC-AI-2024A06-060.

\newpage
\bibliography{custom}

@article{liu2024visual,
  title={Visual instruction tuning},
  author={Liu, Haotian and Li, Chunyuan and Wu, Qingyang and Lee, Yong Jae},
  journal={Advances in neural information processing systems},
  volume={36},
  year={2024}
}

@inproceedings{
    dai2023instructblip,
    title={Instruct{BLIP}: Towards General-purpose Vision-Language Models with Instruction Tuning},
    author={Wenliang Dai and Junnan Li and Dongxu Li and Anthony Tiong and Junqi Zhao and Weisheng Wang and Boyang Li and Pascale Fung and Steven Hoi},
    booktitle={Thirty-seventh Conference on Neural Information Processing Systems},
    year={2023},
    url={https://openreview.net/forum?id=vvoWPYqZJA}
}

@article{tong2024cambrian,
  title={Cambrian-1: A Fully Open, Vision-Centric Exploration of Multimodal LLMs},
  author={Shengbang Tong and Ellis Brown and Penghao Wu and Sanghyun Woo and Manoj Middepogu and Sai Akula and Jihan Yang and Shusheng Yang and Adithya Iyer and Xichen Pan and Austin Wang and Rob Fergus and Yann LeCun and Saining Xie},
  journal={ArXiv},
  year={2024},
  volume={abs/2406.16860},
  url={https://api.semanticscholar.org/CorpusID:270703300}
}

@article{zhou2024lima,
  title={LIMA: Less Is More for Alignment},
  author={Chunting Zhou and Pengfei Liu and Puxin Xu and Srini Iyer and Jiao Sun and Yuning Mao and Xuezhe Ma and Avia Efrat and Ping Yu and L. Yu and Susan Zhang and Gargi Ghosh and Mike Lewis and Luke Zettlemoyer and Omer Levy},
  journal={ArXiv},
  year={2023},
  volume={abs/2305.11206},
  url={https://api.semanticscholar.org/CorpusID:258822910}
}

@misc{chen2023maybe05dataneeded,
      title={Maybe Only 0.5\% Data is Needed: A Preliminary Exploration of Low Training Data Instruction Tuning}, 
      author={Hao Chen and Yiming Zhang and Qi Zhang and Hantao Yang and Xiaomeng Hu and Xuetao Ma and Yifan Yanggong and Junbo Zhao},
      year={2023},
      eprint={2305.09246},
      archivePrefix={arXiv},
      primaryClass={cs.AI}
}

@misc{xia2024lessselectinginfluentialdata,
      title={LESS: Selecting Influential Data for Targeted Instruction Tuning}, 
      author={Mengzhou Xia and Sadhika Malladi and Suchin Gururangan and Sanjeev Arora and Danqi Chen},
      year={2024},
      eprint={2402.04333},
      archivePrefix={arXiv},
      primaryClass={cs.CL},
      url={https://arxiv.org/abs/2402.04333}
}

@article{xue2024xgen,
  title={BLIP-3: A Family of Open Large Multimodal Models},
  author={Le Xue and Manli Shu and Anas Awadalla and Jun Wang and An Yan and Senthil Purushwalkam and Honglu Zhou and Viraj Prabhu and Yutong Dai and Michael S. Ryoo and Shrikant Kendre and Jieyu Zhang and Can Qin and Shu Fang Zhang and Chia-Chih Chen and Ning Yu and Juntao Tan and Tulika Manoj Awalgaonkar and Shelby Heinecke and Huan Wang and Yejin Choi and Ludwig Schmidt and Zeyuan Chen and Silvio Savarese and Juan Carlos Niebles and Caiming Xiong and Ran Xu},
  journal={2025 IEEE/CVF International Conference on Computer Vision Workshops (ICCVW)},
  year={2024},
  pages={6183-6194},
  url={https://api.semanticscholar.org/CorpusID:271891872}
}

@article{chen2025expandingperformanceboundariesopensource,
  title={Expanding Performance Boundaries of Open-Source Multimodal Models with Model, Data, and Test-Time Scaling},
  author={Zhe Chen and Weiyun Wang and Yue Cao and Yangzhou Liu and Zhangwei Gao and Erfei Cui and Jinguo Zhu and Shenglong Ye and Hao Tian and Zhaoyang Liu and Lixin Gu and Xuehui Wang and Qingyun Li and Yiming Ren and Zixuan Chen and Jiapeng Luo and Jiahao Wang and Tan Jiang and Bo Wang and Conghui He and Botian Shi and Xingcheng Zhang and Han Lv and Yi Wang and Wenqi Shao and Pei Chu and Zhongying Tu and Tong He and Zhiyong Wu and Hui Deng and Jiaye Ge and Kaiming Chen and Min Dou and Lewei Lu and Xizhou Zhu and Tong Lu and Dahu Lin and Yunfeng Qiao and Jifeng Dai and Wenhai Wang},
  journal={ArXiv},
  year={2024},
  volume={abs/2412.05271},
  url={https://api.semanticscholar.org/CorpusID:274581884}
}

@misc{li2024llavaonevisioneasyvisualtask,
      title={LLaVA-OneVision: Easy Visual Task Transfer}, 
      author={Bo Li and Yuanhan Zhang and Dong Guo and Renrui Zhang and Feng Li and Hao Zhang and Kaichen Zhang and Peiyuan Zhang and Yanwei Li and Ziwei Liu and Chunyuan Li},
      year={2024},
      eprint={2408.03326},
      archivePrefix={arXiv},
      primaryClass={cs.CV},
      url={https://arxiv.org/abs/2408.03326}
}

@misc{liu2024morehighvaluedataselection,
      title={Less is More: High-value Data Selection for Visual Instruction Tuning}, 
      author={Zikang Liu and Kun Zhou and Wayne Xin Zhao and Dawei Gao and Yaliang Li and Ji-Rong Wen},
      year={2024},
      eprint={2403.09559},
      archivePrefix={arXiv},
      primaryClass={cs.CL},
      url={https://arxiv.org/abs/2403.09559}
}

@article{yu2024mastering,
  title={Mastering Collaborative Multi-modal Data Selection: A Focus on Informativeness, Uniqueness, and Representativeness},
  author={Yu, Qifan and Shen, Zhebei and Yue, Zhongqi and Wu, Yang and Zhang, Wenqiao and Li, Yunfei and Li, Juncheng and Tang, Siliang and Zhuang, Yueting},
  journal={arXiv preprint arXiv:2412.06293},
  year={2024}
}

@misc{lee2024conceptskilltransferabilitybaseddataselection,
      title={Concept-skill Transferability-based Data Selection for Large Vision-Language Models}, 
      author={Jaewoo Lee and Boyang Li and Sung Ju Hwang},
      year={2024},
      eprint={2406.10995},
      archivePrefix={arXiv},
      primaryClass={cs.CV},
      url={https://arxiv.org/abs/2406.10995}
}

@misc{wu2025iconsinfluenceconsensusvisionlanguage,
      title={ICONS: Influence Consensus for Vision-Language Data Selection}, 
      author={Xindi Wu and Mengzhou Xia and Rulin Shao and Zhiwei Deng and Pang Wei Koh and Olga Russakovsky},
      year={2025},
      eprint={2501.00654},
      archivePrefix={arXiv},
      primaryClass={cs.CV},
      url={https://arxiv.org/abs/2501.00654}, 
}

@misc{chen2024visionlanguagemodelstrongfilter,
      title={Your Vision-Language Model Itself Is a Strong Filter: Towards High-Quality Instruction Tuning with Data Selection}, 
      author={Ruibo Chen and Yihan Wu and Lichang Chen and Guodong Liu and Qi He and Tianyi Xiong and Chenxi Liu and Junfeng Guo and Heng Huang},
      year={2024},
      eprint={2402.12501},
      archivePrefix={arXiv},
      primaryClass={cs.CL},
      url={https://arxiv.org/abs/2402.12501}
}

@misc{wei2023instructiongpt4200instructionparadigmfinetuning,
      title={InstructionGPT-4: A 200-Instruction Paradigm for Fine-Tuning MiniGPT-4}, 
      author={Lai Wei and Zihao Jiang and Weiran Huang and Lichao Sun},
      year={2023},
      eprint={2308.12067},
      archivePrefix={arXiv},
      primaryClass={cs.LG},
      url={https://arxiv.org/abs/2308.12067}
}

@misc{zhu2023minigpt4enhancingvisionlanguageunderstanding,
      title={MiniGPT-4: Enhancing Vision-Language Understanding with Advanced Large Language Models}, 
      author={Deyao Zhu and Jun Chen and Xiaoqian Shen and Xiang Li and Mohamed Elhoseiny},
      year={2023},
      eprint={2304.10592},
      archivePrefix={arXiv},
      primaryClass={cs.CV},
      url={https://arxiv.org/abs/2304.10592}
}

@misc{paul2023deeplearningdatadiet,
      title={Deep Learning on a Data Diet: Finding Important Examples Early in Training}, 
      author={Mansheej Paul and Surya Ganguli and Gintare Karolina Dziugaite},
      year={2023},
      eprint={2107.07075},
      archivePrefix={arXiv},
      primaryClass={cs.LG},
      url={https://arxiv.org/abs/2107.07075}
}

@misc{liu2024improvedbaselinesvisualinstruction,
      title={Improved Baselines with Visual Instruction Tuning}, 
      author={Haotian Liu and Chunyuan Li and Yuheng Li and Yong Jae Lee},
      year={2024},
      eprint={2310.03744},
      archivePrefix={arXiv},
      primaryClass={cs.CV},
      url={https://arxiv.org/abs/2310.03744}
}

@inproceedings{xu-etal-2024-vision,
    title = "Vision-Flan: Scaling Human-Labeled Tasks in Visual Instruction Tuning",
    author = "Xu, Zhiyang  and
      Feng, Chao  and
      Shao, Rulin  and
      Ashby, Trevor  and
      Shen, Ying  and
      Jin, Di  and
      Cheng, Yu  and
      Wang, Qifan  and
      Huang, Lifu",
    editor = "Ku, Lun-Wei  and
      Martins, Andre  and
      Srikumar, Vivek",
    booktitle = "Findings of the Association for Computational Linguistics: ACL 2024",
    month = aug,
    year = "2024",
    address = "Bangkok, Thailand",
    publisher = "Association for Computational Linguistics",
    url = "https://aclanthology.org/2024.findings-acl.905/",
    doi = "10.18653/v1/2024.findings-acl.905",
    pages = "15271--15342"
}

@misc{yin2023survey,
      title={MME: A Comprehensive Evaluation Benchmark for Multimodal Large Language Models}, 
      author={Chaoyou Fu and Peixian Chen and Yunhang Shen and Yulei Qin and Mengdan Zhang and Xu Lin and Jinrui Yang and Xiawu Zheng and Ke Li and Xing Sun and Yunsheng Wu and Rongrong Ji and Caifeng Shan and Ran He},
      year={2025},
      eprint={2306.13394},
      archivePrefix={arXiv},
      primaryClass={cs.CV},
      url={https://arxiv.org/abs/2306.13394}, 
}

@misc{liu2024mmbenchmultimodalmodelallaround,
    title={MMBench: Is Your Multi-modal Model an All-around Player?},
    author={Yuan Liu and Haodong Duan and Yuanhan Zhang and Bo Li and Songyang Zhang and Wangbo Zhao and Yike Yuan and Jiaqi Wang and Conghui He and Ziwei Liu and Kai Chen and Dahua Lin},
    year={2024},
    eprint={2307.06281},
    archivePrefix={arXiv},
    primaryClass={[cs.CV](http://cs.cv/)},
    url={https://arxiv.org/abs/2307.06281}
}

@inproceedings{
lu2022learn,
title={Learn to Explain: Multimodal Reasoning via Thought Chains for Science Question Answering},
author={Pan Lu and Swaroop Mishra and Tony Xia and Liang Qiu and Kai-Wei Chang and Song-Chun Zhu and Oyvind Tafjord and Peter Clark and Ashwin Kalyan},
booktitle={Advances in Neural Information Processing Systems},
editor={Alice H. Oh and Alekh Agarwal and Danielle Belgrave and Kyunghyun Cho},
year={2022},
url={https://openreview.net/forum?id=HjwK-Tc_Bc}
}

@misc{li2023evaluatingobjecthallucinationlarge,
      title={Evaluating Object Hallucination in Large Vision-Language Models}, 
      author={Yifan Li and Yifan Du and Kun Zhou and Jinpeng Wang and Wayne Xin Zhao and Ji-Rong Wen},
      year={2023},
      eprint={2305.10355},
      archivePrefix={arXiv},
      primaryClass={cs.CV},
      url={https://arxiv.org/abs/2305.10355}
}

@misc{gurari2018vizwizgrandchallengeanswering,
      title={VizWiz Grand Challenge: Answering Visual Questions from Blind People}, 
      author={Danna Gurari and Qing Li and Abigale J. Stangl and Anhong Guo and Chi Lin and Kristen Grauman and Jiebo Luo and Jeffrey P. Bigham},
      year={2018},
      eprint={1802.08218},
      archivePrefix={arXiv},
      primaryClass={cs.CV},
      url={https://arxiv.org/abs/1802.08218}
}

@misc{yu2024mmvetevaluatinglargemultimodal,
      title={MM-Vet: Evaluating Large Multimodal Models for Integrated Capabilities}, 
      author={Weihao Yu and Zhengyuan Yang and Linjie Li and Jianfeng Wang and Kevin Lin and Zicheng Liu and Xinchao Wang and Lijuan Wang},
      year={2024},
      eprint={2308.02490},
      archivePrefix={arXiv},
      primaryClass={cs.AI},
      url={https://arxiv.org/abs/2308.02490}
}

@article{yue2024mmmumassivemultidisciplinemultimodal,
  title={MMMU: A Massive Multi-Discipline Multimodal Understanding and Reasoning Benchmark for Expert AGI},
  author={Xiang Yue and Yuansheng Ni and Kai Zhang and Tianyu Zheng and Ruoqi Liu and Ge Zhang and Samuel Stevens and Dongfu Jiang and Weiming Ren and Yuxuan Sun and Cong Wei and Botao Yu and Ruibin Yuan and Renliang Sun and Ming Yin and Boyuan Zheng and Zhenzhu Yang and Yibo Liu and Wenhao Huang and Huan Sun and Yu Su and Wenhu Chen},
  journal={2024 IEEE/CVF Conference on Computer Vision and Pattern Recognition (CVPR)},
  year={2023},
  pages={9556-9567},
  url={https://api.semanticscholar.org/CorpusID:265466525}
}

@misc{zhang2024wingslearningmultimodalllms,
      title={Wings: Learning Multimodal LLMs without Text-only Forgetting}, 
      author={Yi-Kai Zhang and Shiyin Lu and Yang Li and Yanqing Ma and Qing-Guo Chen and Zhao Xu and Weihua Luo and Kaifu Zhang and De-Chuan Zhan and Han-Jia Ye},
      year={2024},
      eprint={2406.03496},
      archivePrefix={arXiv},
      primaryClass={cs.CL},
      url={https://arxiv.org/abs/2406.03496}
}

@article{levi2024double,
  title={The Double-Ellipsoid Geometry of CLIP},
  author={Levi, Meir Yossef and Gilboa, Guy},
  journal={arXiv preprint arXiv:2411.14517},
  year={2024}
}

@misc{gao2019representationdegenerationproblemtrainingarxiv,
      title={Representation Degeneration Problem in Training Natural Language Generation Models}, 
      author={Jun Gao and Di He and Xu Tan and Tao Qin and Liwei Wang and Tie-Yan Liu},
      year={2019},
      eprint={1907.12009},
      archivePrefix={arXiv},
      primaryClass={cs.CL},
      url={https://arxiv.org/abs/1907.12009}, 
}

@inproceedings{zhang-etal-2020-unsupervised,
    title = "An Unsupervised Sentence Embedding Method by Mutual Information Maximization",
    author = "Zhang, Yan  and
      He, Ruidan  and
      Liu, Zuozhu  and
      Lim, Kwan Hui  and
      Bing, Lidong",
    editor = "Webber, Bonnie  and
      Cohn, Trevor  and
      He, Yulan  and
      Liu, Yang",
    booktitle = "Proceedings of the 2020 Conference on Empirical Methods in Natural Language Processing (EMNLP)",
    month = nov,
    year = "2020",
    address = "Online",
    publisher = "Association for Computational Linguistics",
    url = "https://aclanthology.org/2020.emnlp-main.124/",
    doi = "10.18653/v1/2020.emnlp-main.124",
    pages = "1601--1610"
}

@inproceedings{radford2021learning,
    title={Learning Transferable Visual Models From Natural Language Supervision},
    author={Alec Radford and Jong Wook Kim and Chris Hallacy and Aditya Ramesh and Gabriel Goh and Sandhini Agarwal and Girish Sastry and Amanda Askell and Pamela Mishkin and Jack Clark and Gretchen Krueger and Ilya Sutskever},
    booktitle={Proceedings of the 38th International Conference on Machine Learning (ICML)},
    year={2021},
    pages={8748--8763},
    volume={139},
}

@inproceedings{ethayarajh-2019-contextual,
    title = "How Contextual are Contextualized Word Representations? {C}omparing the Geometry of {BERT}, {ELM}o, and {GPT}-2 Embeddings",
    author = "Ethayarajh, Kawin",
    booktitle = "Proceedings of the 2019 Conference on Empirical Methods in Natural Language Processing and the 9th International Joint Conference on Natural Language Processing (EMNLP-IJCNLP)",
    month = nov,
    year = "2019",
    address = "Hong Kong, China",
    publisher = "Association for Computational Linguistics",
    url = "https://www.aclweb.org/anthology/D19-1006",
    doi = "10.18653/v1/D19-1006",
    pages = "55--65",
}

@article{arora-etal-2016-latent,
    title = "A Latent Variable Model Approach to {PMI}-based Word Embeddings",
    author = "Arora, Sanjeev  and
      Li, Yuanzhi  and
      Liang, Yingyu  and
      Ma, Tengyu  and
      Risteski, Andrej",
    editor = "Lee, Lillian  and
      Johnson, Mark  and
      Toutanova, Kristina",
    journal = "Transactions of the Association for Computational Linguistics",
    volume = "4",
    year = "2016",
    address = "Cambridge, MA",
    publisher = "MIT Press",
    url = "https://aclanthology.org/Q16-1028/",
    doi = "10.1162/tacl_a_00106",
    pages = "385--399"
}

@misc{hu2021lora,
      title={{LoRA}: Low-Rank Adaptation of Large Language Models}, 
      author={Edward J. Hu and Yelong Shen and Phillip Wallis and Zeyuan Allen-Zhu and Yuanzhi Li and Shean Wang and Lu Wang and Weizhu Chen},
      year={2021},
      eprint={2106.09685},
      archivePrefix={arXiv},
      primaryClass={cs.CL}
}

@article{jing2021understanding,
  title={Understanding dimensional collapse in contrastive self-supervised learning},
  author={Jing, Li and Vincent, Pascal and LeCun, Yann and Tian, Yuandong},
  journal={arXiv preprint arXiv:2110.09348},
  year={2021}
}

@inproceedings{bi-etal-2025-llava,
    title = "{LL}a{VA} Steering: Visual Instruction Tuning with 500x Fewer Parameters through Modality Linear Representation-Steering",
    author = "Bi, Jinhe  and
      Wang, Yujun  and
      Chen, Haokun  and
      Xiao, Xun  and
      Hecker, Artur  and
      Tresp, Volker  and
      Ma, Yunpu",
    editor = "Che, Wanxiang  and
      Nabende, Joyce  and
      Shutova, Ekaterina  and
      Pilehvar, Mohammad Taher",
    booktitle = "Proceedings of the 63rd Annual Meeting of the Association for Computational Linguistics (Volume 1: Long Papers)",
    month = jul,
    year = "2025",
    address = "Vienna, Austria",
    publisher = "Association for Computational Linguistics",
    url = "https://aclanthology.org/2025.acl-long.739/",
    pages = "15230--15250",
    ISBN = "979-8-89176-251-0"
}

@article{bi2025cot,
  title={CoT-Kinetics: A Theoretical Modeling Assessing LRM Reasoning Process},
  author={Jinhe Bi and Danqi Yan and Yifan Wang and Wenke Huang and Haokun Chen and Guancheng Wan and Mang Ye and Xun Xiao and Hin-rich Schuetze and Volker Tresp and Yunpu Ma},
  journal={ArXiv},
  year={2025},
  volume={abs/2505.13408},
  url={https://api.semanticscholar.org/CorpusID:278769227}
}

@misc{lu2025mvdebatemultiviewagentdebate,
      title={MV-Debate: Multi-view Agent Debate with Dynamic Reflection Gating for Multimodal Harmful Content Detection in Social Media}, 
      author={Rui Lu and Jinhe Bi and Yunpu Ma and Feng Xiao and Yuntao Du and Yijun Tian},
      year={2025},
      eprint={2508.05557},
      archivePrefix={arXiv},
      primaryClass={cs.AI},
      url={https://arxiv.org/abs/2508.05557}, 
}

@misc{tian2025reinforcementmidtraining,
      title={Reinforcement Mid-Training}, 
      author={Yijun Tian and Shaoyu Chen and Zhichao Xu and Yawei Wang and Jinhe Bi and Peng Han and Wei Wang},
      year={2025},
      eprint={2509.24375},
      archivePrefix={arXiv},
      primaryClass={cs.CL},
      url={https://arxiv.org/abs/2509.24375}, 
}

@misc{peng2025visualinputcompressedvisual,
      title={Can Visual Input Be Compressed? A Visual Token Compression Benchmark for Large Multimodal Models}, 
      author={Tianfan Peng and Yuntao Du and Pengzhou Ji and Shijie Dong and Kailin Jiang and Mingchuan Ma and Yijun Tian and Jinhe Bi and Qian Li and Wei Du and Feng Xiao and Lizhen Cui},
      year={2025},
      eprint={2511.02650},
      archivePrefix={arXiv},
      primaryClass={cs.CV},
      url={https://arxiv.org/abs/2511.02650}, 
}

@misc{huang2025loongsynthesizelongchainofthoughts,
      title={Loong: Synthesize Long Chain-of-Thoughts at Scale through Verifiers}, 
      author={Xingyue Huang and Rishabh and Gregor Franke and Ziyi Yang and Jiamu Bai and Weijie Bai and Jinhe Bi and Zifeng Ding and Yiqun Duan and Chengyu Fan and Wendong Fan and Xin Gao and Ruohao Guo and Yuan He and Zhuangzhuang He and Xianglong Hu and Neil Johnson and Bowen Li and Fangru Lin and Siyu Lin and Tong Liu and Yunpu Ma and Hao Shen and Hao Sun and Beibei Wang and Fangyijie Wang and Hao Wang and Haoran Wang and Yang Wang and Yifeng Wang and Zhaowei Wang and Ziyang Wang and Yifan Wu and Zikai Xiao and Chengxing Xie and Fan Yang and Junxiao Yang and Qianshuo Ye and Ziyu Ye and Guangtao Zeng and Yuwen Ebony Zhang and Zeyu Zhang and Zihao Zhu and Bernard Ghanem and Philip Torr and Guohao Li},
      year={2025},
      eprint={2509.03059},
      archivePrefix={arXiv},
      primaryClass={cs.LG},
      url={https://arxiv.org/abs/2509.03059}, 
}

@misc{yang2026alignsaeconceptalignedsparseautoencoders,
      title={AlignSAE: Concept-Aligned Sparse Autoencoders}, 
      author={Minglai Yang and Xinyu Guo and Zhengliang Shi and Jinhe Bi and Steven Bethard and Mihai Surdeanu and Liangming Pan},
      year={2026},
      eprint={2512.02004},
      archivePrefix={arXiv},
      primaryClass={cs.LG},
      url={https://arxiv.org/abs/2512.02004}, 
}

@misc{wan2025magicwordssharpnessawareprompt,
      title={Beyond Magic Words: Sharpness-Aware Prompt Evolving for Robust Large Language Models with TARE}, 
      author={Guancheng Wan and Lucheng Fu and Haoxin Liu and Yiqiao Jin and Hui Yi Leong and Eric Hanchen Jiang and Hejia Geng and Jinhe Bi and Yunpu Ma and Xiangru Tang and B. Aditya Prakash and Yizhou Sun and Wei Wang},
      year={2025},
      eprint={2509.24130},
      archivePrefix={arXiv},
      primaryClass={cs.CL},
      url={https://arxiv.org/abs/2509.24130}, 
}

@misc{li2026graphsubstratedatamodalities,
      title={Graph is a Substrate Across Data Modalities}, 
      author={Ziming Li and Xiaoming Wu and Zehong Wang and Jiazheng Li and Yijun Tian and Jinhe Bi and Yunpu Ma and Yanfang Ye and Chuxu Zhang},
      year={2026},
      eprint={2601.22384},
      archivePrefix={arXiv},
      primaryClass={cs.LG},
      url={https://arxiv.org/abs/2601.22384}, 
}

@inproceedings{
wan2025hyperion,
title={{HYPERION}: Fine-Grained Hypersphere Alignment for Robust Federated Graph Learning},
author={Guancheng Wan and Xiaoran Shang and Yuxin Wu and Guibin Zhang and Jinhe Bi and Liangtao Zheng and Xin Lin and Yue Liu and Yanbiao Ma and Wenke Huang and Bo Du},
booktitle={The Thirty-ninth Annual Conference on Neural Information Processing Systems},
year={2025},
url={https://openreview.net/forum?id=TZB6YT8Owr}
}

@misc{bi2026reflectrllearninggoldennegative,
      title={ReflectRL: Learning from Golden Negative Trajectories via Reflective-to-Direct Reasoning}, 
      author={Jinhe Bi and Chennan Zhou and Zengjie Jin and Aniri and Shuo Lu and Wenke Huang and Hu Cao and Xun Xiao and Zhihong Zhu and Volker Tresp and Fei Shen and Yunpu Ma and Tat-Seng Chua},
      year={2026},
      eprint={2608.03972},
      archivePrefix={arXiv},
      primaryClass={cs.AI},
      url={https://arxiv.org/abs/2608.03972}, 
}

@inproceedings{
bi2026echorl,
title={Echo{RL}: Reinforcement Learning via Rollout Echoing},
author = {Jinhe Bi and Aniri and Minglai Yang and Xingcheng Zhou and Wenke Huang and Sikuan Yan and Yujun Wang and Zixuan Cao and Michael Färber and Xun Xiao and Volker Tresp and Yunpu Ma}
,
booktitle={Forty-third International Conference on Machine Learning},
year={2026},
url={https://openreview.net/forum?id=A6az59SGtF}
}

@misc{aniri2026opdvvisualonpolicyselfdistillation,
      title={OPD-V: Visual On-Policy Self-Distillation with Modality Balance}, 
      author={Aniri and Jinhe Bi and Peng Liao and Zengjie Jin and Volker Tresp and Fei Shen and Yunpu Ma and Tat-Seng Chua},
      year={2026},
      eprint={2608.05131},
      archivePrefix={arXiv},
      primaryClass={cs.CV},
      url={https://arxiv.org/abs/2608.05131}, 
}

@misc{aniri2026fabrimaeitrustmyself,
      title={FabriMAE I Trust Myself? Self-Evaluating VLA Action Generation with Markov Attention Entropy}, 
      author={Aniri and Chen Yilin and Jinhe Bi and Junfei Guo and Donglai Ran and Xu Bian and Zengjie Jin and Yujun Wang and Yijun Tian and Volker Tresp and Fei Shen and Tat-Seng Chua and Yunpu Ma},
      year={2026},
      eprint={2608.16697},
      archivePrefix={arXiv},
      primaryClass={cs.AI},
      url={https://arxiv.org/abs/2608.16697}, 
}

\clearpage
\appendix
\section{Related Work}

\begin{figure*}[ht]
    \centering
    \begin{minipage}{0.48\textwidth}
        \centering
        \includegraphics[width=\linewidth]{figs/COCO_train2014_000000344067.jpg}
        \textbf{Top-Ranked Sample} \\
        \begin{flushleft}
        \small
        \textbf{[Question]} What region of the world is this located? \\
        \textbf{[Answer]} Asia \\
        \textbf{[Reasoning]} Requires fine-grained OCR and cultural reasoning.
        \end{flushleft}
    \end{minipage}
    \hfill
    \begin{minipage}{0.43\textwidth}
        \centering
        \includegraphics[width=\linewidth]{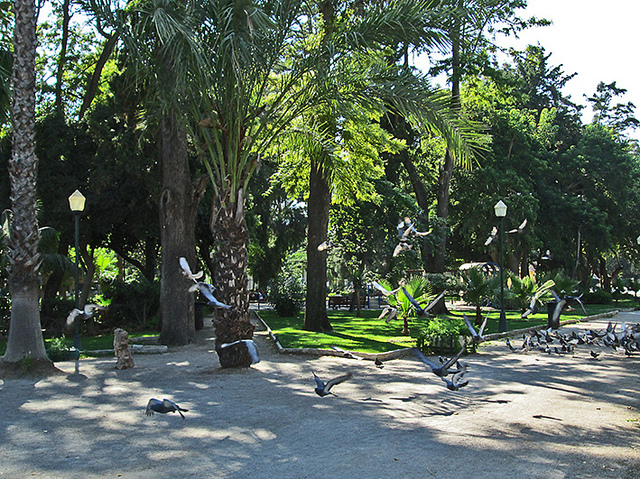}
        \textbf{Bottom-Ranked Sample} \\
        \begin{flushleft}
        \small
        \textbf{[Question]} What type of weather is this? \\
        \textbf{[Answer]} Sunny \\
        \textbf{[Reasoning]} Trivial perception; high semantic redundancy.
        \end{flushleft}
    \end{minipage}
    \caption{\textbf{Representative examples selected by PRISM from LLaVA-665K.} Top-ranked samples correspond to more challenging and informative interactions that exhibit unique semantic structures, whereas bottom-ranked samples tend to be trivial, binary, or semantically redundant.}
    \label{fig:case_study}
\end{figure*}

\textbf{Visual Instruction Tuning:}  
Visual instruction tuning is essential for aligning MLLMs with both practical applications and academic benchmarks. Early methods relied on synthetic visual instructions, which performed well in conversations but struggled on rigorous benchmarks. A hybrid approach later emerged, integrating synthetic data with academic datasets to improve training diversity. This advancement has enhanced models like LLaVA \citep{liu2024visual}, InstructBLIP \citep{dai2023instructblip}, and Cambrian \citep{tong2024cambrian}, enabling better visual-linguistic understanding. Beyond task performance, visual instruction tuning improves model alignment with user expectations, ensuring both practical utility and strong academic performance.

\noindent
\textbf{Visual Instruction Selection:}  
Despite the strong performance of MLLMs, the rapid growth of visual instruction datasets has introduced significant redundancy, similar to challenges in LLMs \citep{zhou2024lima,chen2023maybe05dataneeded,xia2024lessselectinginfluentialdata,tian2025reinforcementmidtraining,lu2025mvdebatemultiviewagentdebate,bi2025cot,peng2025visualinputcompressedvisual,wan2025hyperion,wan2025magicwordssharpnessawareprompt,huang2025loongsynthesizelongchainofthoughts,li2026graphsubstratedatamodalities,yang2026alignsaeconceptalignedsparseautoencoders,bi2026echorl,bi2026reflectrllearninggoldennegative}. State-of-the-art models like BLIP3 \citep{xue2024xgen}, InternVL2.5 \citep{chen2025expandingperformanceboundariesopensource}, and LLaVA-OneVision \citep{li2024llavaonevisioneasyvisualtask} rely on billions of instructions to enhance understanding, but their massive scale leads to substantial computational costs, often requiring hundreds to thousands of GPU hours.

\noindent To address this, various data selection strategies aim to reduce redundancy while preserving performance. TIVE \citep{liu2024morehighvaluedataselection} selects valuable data based on gradient similarity but requires additional training on downstream tasks. SELF-FILTER \citep{chen2024visionlanguagemodelstrongfilter} uses an auxiliary evaluation model to prioritize high-value samples. COINCIDE \citep{lee2024conceptskilltransferabilitybaseddataselection} clusters data by conceptual and skill-based representations, while InstructionGPT-4 \citep{wei2023instructiongpt4200instructionparadigmfinetuning} filters 200 instructions for MiniGPT-4 \citep{zhu2023minigpt4enhancingvisionlanguageunderstanding}, though it lacks scalability. ICONS \citep{wu2025iconsinfluenceconsensusvisionlanguage} extends LESS \citep{xia2024lessselectinginfluentialdata} by incorporating specialist influence estimation for instruction tuning. DataTailor \citep{yu2024mastering} selects data based on informativeness, uniqueness, and representativeness to retain the most relevant samples.
\section{Dataset and Selection Details}
\label{sec:appendix_dataset}

This section provides a detailed breakdown of our \texttt{PRISM-Instruct-250K} dataset and the underlying data distributions that informed its creation. The dataset is the result of applying the PRISM selection methodology to the LLaVA-665K data pool, which aggregates samples from diverse sources. Our goal is to create a compact, high-utility subset by retaining the most informative, low-redundancy samples.

\subsection{Redundancy Score Distribution}
\begin{figure}[ht!]
    \centering
    \includegraphics[width=1\linewidth]{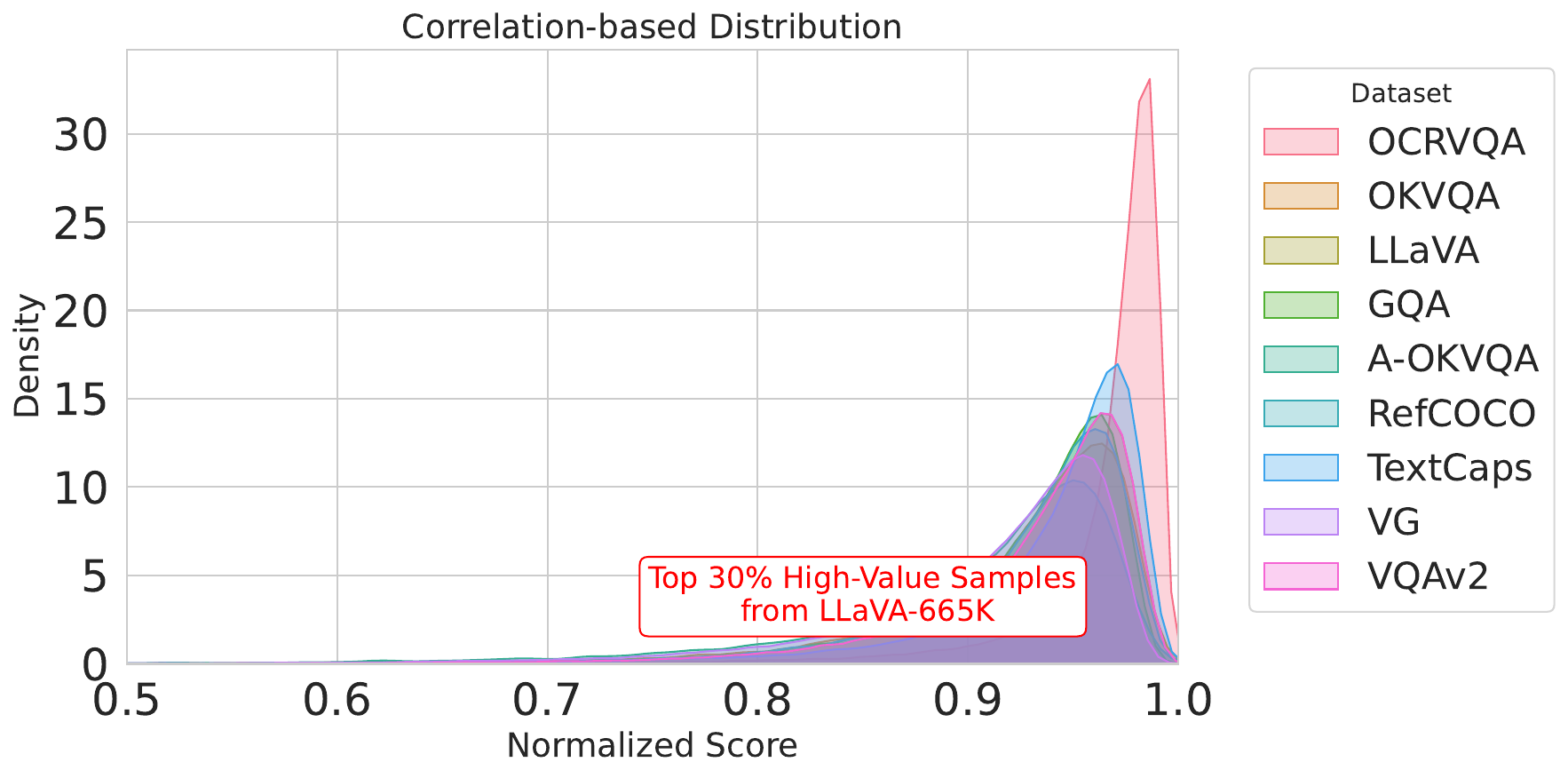}
    \caption{
        \textbf{Distribution of PRISM's Redundancy Scores across source datasets.} The vertical line indicates the global selection threshold used to prune the 70\% most redundant samples (i.e., those with the highest correlation scores). Datasets with a higher density of low-correlation samples contribute more instances to the final \texttt{PRISM-Instruct-250K} set, ensuring a merit-based composition.
    }
    \label{fig:score_distribution}
\end{figure}

At the core of our selection process is the \textit{Redundancy Score} $\mathcal{R}(d_i)$, which quantifies the degree to which a sample is semantically similar to the rest of the corpus. Figure~\ref{fig:score_distribution} visualizes the distribution of these scores (lower is better) for each source dataset within the LLaVA-665K pool.

The distributions reveal the varying levels of internal diversity within each source. For instance, datasets like LLaVA and VG exhibit a wider distribution with a significant number of low-score (high-value) samples. In contrast, other datasets may have a higher density of redundant, high-score samples. PRISM applies a single, global threshold to this combined pool, ensuring that only the top 30\% of samples—those with the lowest redundancy scores, regardless of their origin—are selected. This principled approach prevents any single, potentially redundant dataset from dominating the final selection and instead curates a subset based on universal informational value.

\paragraph{Scope of re-centering.}
We emphasize that PRISM does not claim to perform full whitening or to make the empirical covariance matrix exactly proportional to the identity. Instead, PRISM targets the dominant first-order component of visual representation anisotropy: the corpus-level mean shift. Our analysis shows that this global component can dominate raw cosine similarity and obscure sample-specific semantic variation. Therefore, the goal of PRISM is not to completely isotropize the representation space, but to remove the shared global drift that is most harmful for redundancy estimation.

\paragraph{Why mean re-centering is sufficient for selection.}
For the purpose of training-free data selection, full covariance correction is not always necessary. Redundancy ranking only requires a similarity signal that better reflects relative sample-specific variation. By subtracting the corpus mean and computing normalized inner products in the centered space, PRISM suppresses the common background direction while preserving an efficient one-pass scoring pipeline. This design deliberately trades complete second-order correction for scalability, numerical stability, and ease of deployment on large multimodal instruction pools.

\paragraph{Why not full whitening or principal-component removal?}
A natural alternative is to explicitly whiten the representation space or remove the top principal components. We view these methods as complementary but less aligned with our objective. First, whitening requires reliable covariance estimation in a high-dimensional, low-effective-rank regime, where the covariance matrix can be ill-conditioned and requires additional regularization choices. Second, top-PC removal introduces an extra hyperparameter: the number of components to remove. This hyperparameter can be dataset- and model-dependent, reducing the plug-and-play nature of training-free selection. Third, both approaches increase implementation complexity compared with the corpus-mean re-centering used by PRISM. PRISM therefore focuses on the first-order global drift, which yields a strong efficiency--performance trade-off without additional tuning.

\begin{table*}[t]
\centering
\large
\label{tab:geometry_baseline_applicability}
\resizebox{\linewidth}{!}{
\begin{tabular}{lccccp{6.8cm}}
\toprule
\textbf{Method} &
\textbf{Feature Space} &
\textbf{Corrects Mean Drift?} &
\textbf{Extra Hyperparam.?} &
\textbf{Scalable?} &
\textbf{Role / Limitation} \\
\midrule
Random selection &
None &
No &
No &
Yes &
Matched-budget lower-bound control. It tests whether gains come merely from using fewer samples, but it does not exploit visual geometry or redundancy. \\

Raw cosine redundancy &
Target MLLM features &
No &
No &
Yes &
Direct ablation of PRISM's re-centering. It remains dominated by the corpus-level drift in anisotropic visual spaces and therefore measures shared background directions rather than sample-specific semantic variation. \\

PRISM&
Target MLLM features &
Yes &
No &
Yes &
Our default intrinsic selector. It removes the first-order corpus-level shift and computes redundancy through an exact aggregate form without constructing an $N\times N$ similarity matrix. \\

Top-PC removal + cosine &
Target MLLM features &
Partially &
Yes: number of PCs $k$ &
Moderate &
A reasonable diagnostic alternative, but it introduces a dataset/model-dependent choice of $k$. Removing too few PCs may leave drift; removing too many may discard semantic directions useful for instruction tuning. \\

Whitening / Mahalanobis-style correction &
Target MLLM features &
Yes, plus covariance correction &
Yes: shrinkage / rank / clipping &
No / costly &
More aggressive second-order correction, but requires estimating and inverting a high-dimensional covariance matrix. In low-effective-rank visual spaces, this is ill-conditioned and requires regularization, weakening plug-and-play training-free deployment. \\

Centered $k$-center / farthest-point sampling &
Target MLLM features &
Yes, if centered &
Yes: distance metric / approximation &
No for exact greedy &
Optimizes geometric coverage rather than corpus-wide redundancy degree. Exact greedy selection is expensive at LLaVA-665K scale; approximate variants introduce additional engineering and approximation choices. \\

CLIP-feature selection &
External / vision-encoder feature space &
Depends on preprocessing &
Possibly &
Yes &
Tests whether generic vision features are sufficient. However, it is not an intrinsic selector over the target MLLM's post-projection representation, so it may be misaligned with the actual instruction-tuning model. \\
\bottomrule
\end{tabular}
}
\caption{
\textbf{Applicability analysis of simple geometry baselines.}
We compare PRISM with natural geometry-based alternatives under the desiderata of training-free multimodal instruction selection:
(i) correcting the dominant global drift, (ii) avoiding additional tuning, (iii) preserving linear or near-linear scalability, and
(iv) operating on the target MLLM's intrinsic feature space. These alternatives are useful diagnostic controls, but most fail at least one desideratum.
}
\end{table*}

\subsection{Final Dataset Composition}
The application of our selection strategy results in the \texttt{PRISM-Instruct-250K} dataset. Table~\ref{tab:prism_data_distribution} details the final number of samples drawn from each source. The composition reflects the data-driven nature of our method: sources rich in unique, low-correlation samples are more heavily represented. The result is a balanced and diverse instruction-tuning dataset that has been algorithmically optimized for both performance and efficiency.

\begin{table}[ht!]
\centering
\small
\begin{tabular}{l r}
\toprule
\textbf{Data Source} & \textbf{Number of Samples} \\
\midrule
LLaVA & 53,591 \\
VQAv2 & 27,567 \\
OKVQA & 2,997 \\
A-OKVQA & 22,032 \\
RefCOCO & 16,933 \\
VG & 28,777 \\
GQA & 24,023 \\
OCRVQA & 26,638 \\
TextCaps & 7,311 \\
Text-Only & 40,688 \\
\midrule
\textbf{Total} & \textbf{250,557} \\
\bottomrule
\end{tabular}
\caption{Final sample distribution of the \texttt{PRISM-Instruct-250K} dataset after applying our selection methodology to the LLaVA-665K pool.}
\label{tab:prism_data_distribution}
\end{table}

\subsection{Language Knowledge Retention Details}
\label{sec:appendix_text_retention}

Table~\ref{tab:text} reports the text-only evaluation underlying Fig.~\ref{fig:flan_llm_summary}. Across three model scales, PRISM preserves or improves language knowledge after visual instruction tuning, with the largest gain observed on the 13B Vicuna backbone.

\begin{table}[ht!]
\centering
\resizebox{\columnwidth}{!}{
\begin{tabular}{l*{4}{c}}
\toprule
\textbf{Model} & \textbf{HellaSwag} & \textbf{MMLU} & \textbf{MMLU-PRO} & \textbf{Rel. (\%)} \\
\midrule
LLaVA-Phi2-3B & 66.0 & 50.5 & 9.1 & 100.0 \\
\cellcolor{cyan!10}\includegraphics[height=1em]{figs/Prismatic_Star.png}PRISM-3B
& \cellcolor{cyan!10}\textbf{67.4}
& \cellcolor{cyan!10}\textbf{52.7}
& \cellcolor{cyan!10}8.6
& \cellcolor{cyan!10}\textbf{100.3} \\
\cmidrule(lr){1-5}
LLaVA-Vicuna-7B & 66.5 & 35.0 & 17.8 & 100.0 \\
\cellcolor{cyan!10}\includegraphics[height=1em]{figs/Prismatic_Star.png}PRISM-7B
& \cellcolor{cyan!10}\textbf{66.5}
& \cellcolor{cyan!10}\textbf{41.1}
& \cellcolor{cyan!10}15.7
& \cellcolor{cyan!10}\textbf{101.9} \\
\cmidrule(lr){1-5}
LLaVA-Vicuna-13B & 69.5 & 36.2 & 6.8 & 100.0 \\
\cellcolor{cyan!10}\includegraphics[height=1em]{figs/Prismatic_Star.png}PRISM-13B
& \cellcolor{cyan!10}\textbf{69.6}
& \cellcolor{cyan!10}\textbf{39.5}
& \cellcolor{cyan!10}\textbf{12.4}
& \cellcolor{cyan!10}\textbf{130.6} \\
\bottomrule
\end{tabular}
}
\caption{Text-only performance after visual instruction tuning. Relative scores are normalized to each full-data model.}
\label{tab:text}
\end{table}

\subsection{Robustness on VisionFlan}
\label{sec:visionflan_robustness}

We further evaluate PRISM on VisionFlan-186K \citep{xu-etal-2024-vision}, a task-diverse instruction pool containing 191 distinct vision-language tasks and 186K samples. Unlike the LLaVA-665K mixture, VisionFlan preserves fine-grained task diversity and therefore provides a complementary test of whether PRISM remains effective under a different instruction distribution.

Using the same 30\% selection budget, PRISM selects 57K samples from VisionFlan-186K. As shown in Table~\ref{tab:visionflan}, full-data training reaches 100.0\% relative performance, random selection with 57K samples reaches 94.1\%, and PRISM reaches 100.9\%. PRISM slightly exceeds the full-data baseline while using only about one third of the instruction pool.

\begin{table*}[ht!]
\centering
\resizebox{\textwidth}{!}{
\begin{tabular}{l*{7}{c}}
\toprule
\textbf{Method} & \textbf{VizWiz} & \textbf{SQA-I} & \textbf{TextVQA} & \textbf{POPE} & \textbf{MME} & \textbf{MMBench} & \textbf{Rel. (\%)} \\
\midrule
Full Data (186K) & 41.7 & 60.8 & 50.4 & 83.4 & 1263.2 & 52.6 & 100.0 \\
Random (57K) & 38.8 & 56.5 & 46.9 & 83.1 & 1175.0 & 48.9 & 94.1 \\
\cellcolor{cyan!10}\includegraphics[height=1em]{figs/Prismatic_Star.png}PRISM (57K)
& \cellcolor{cyan!10}\textbf{42.3}
& \cellcolor{cyan!10}\textbf{61.1}
& \cellcolor{cyan!10}\textbf{50.8}
& \cellcolor{cyan!10}\textbf{84.1}
& \cellcolor{cyan!10}\textbf{1275.5}
& \cellcolor{cyan!10}\textbf{53.1}
& \cellcolor{cyan!10}\textbf{100.9} \\
\bottomrule
\end{tabular}
}
\caption{Robustness evaluation on VisionFlan-186K. PRISM selects 57K samples and matches or exceeds full-data training across the aggregate score.}
\label{tab:visionflan}
\end{table*}
\section{Selection Budget and Relative Score Calculation}
\label{app:budget_rel_score}

\paragraph{Clarification of the 30\% selection budget.}
The 30\% budget used in our main experiments refers to the selection budget over image-associated multimodal samples, not over the entire LLaVA-665K mixture. Following the data composition protocol of LLaVA-1.5, the instruction-tuning mixture contains both visual instruction samples and text-only language samples. Since PRISM is designed to estimate redundancy from visual representations, we apply PRISM only to the visual subset and retain the lowest-redundancy 30\% visual samples. The text-only samples are kept unchanged across PRISM and all comparable baselines.

This design choice has two purposes. First, it isolates the effect of visual-instruction pruning, ensuring that performance differences are not caused by different amounts of language-only supervision. Second, it follows the original LLaVA-1.5 mixture structure, where language-only samples are included to preserve the language capability of the base model. As a result, the final PRISM-Instruct subset contains the selected 30\% visual samples together with the unchanged text-only samples. In our LLaVA-665K setting, this yields 250,557 total training examples, corresponding to approximately 38\% of the full mixture.

\paragraph{Fairness protocol for baseline comparison.}
For a fair comparison, all selection methods should follow the same data-budget rule: the selection budget is applied to the visual subset only, while the text-only subset is retained unchanged. In addition, all methods should use the same base model, visual instruction pool, training recipe, number of epochs, evaluation scripts, and, when applicable, random seed. For baselines reproduced in our pipeline, we enforce this protocol directly. For baselines adopted from prior work, we mark their provenance explicitly and only compare them when their reported setting matches the LLaVA-1.5-7B/LLaVA-665K setup as closely as possible.

\paragraph{Relative score calculation under missing entries.}
To avoid imputing unreported results, we compute the relative score only over the set of available metrics for each method. Specifically, let $\mathcal{M}_s$ denote the set of metrics reported for method $s$, and let $P_{s,m}$ and $P_{\mathrm{full},m}$ denote the score of method $s$ and the full-data fine-tuning baseline on metric $m$, respectively. Since all metrics in our table are higher-is-better, the relative score is computed as
\[
\mathrm{Rel.}(s)
=
100 \times
\frac{1}{|\mathcal{M}_s|}
\sum_{m \in \mathcal{M}_s}
\frac{P_{s,m}}{P_{\mathrm{full},m}}.
\]
For metrics with multiple subsets, such as POPE-P, POPE-R, and POPE-A, each reported subset is treated as one metric. If a prior work only reports the average over POPE subsets, we use the reported average only for that baseline and mark this in the table caption.

Because methods with missing entries are averaged over different metric sets, their relative scores should be interpreted as available-metric relative performance rather than a fully matched aggregate across all benchmarks. Our main claims about PRISM are based on the complete metric set, since PRISM and the full-data baseline are evaluated on all benchmarks under the same protocol.

\section{Model Architecture Details and Generalization Setup}
\label{sec:appendix_model_arch}

To rigorously evaluate the generalization capabilities of PRISM, we extend our analysis across a diverse set of Multimodal Large Language Models (MLLMs). The objective is to verify whether the visual anisotropy diagnosis and the subsequent performance gains from PRISM remain consistent across different model scales, language model families, and vision encoder architectures. This broad evaluation ensures that the intrinsic semantic properties identified by PRISM are not artifacts of a specific model configuration but represent fundamental characteristics of multimodal data representations.

As summarized in Table~\ref{tab:model_arch}, our evaluation framework encompasses the following dimensions:
\begin{itemize}
    \item \textbf{Language Model Families and Scales}: Beyond the Vicuna and Phi-2 families discussed in the main text, we include the Qwen2.5 series (7B Base and Instruct) and Llama-3 (8B) to represent state-of-the-art open-source LLM backbones.
    \item \textbf{Vision Encoders}: We evaluate models utilizing various vision foundation backbones, specifically CLIP ViT-L/14 (336px) and SigLIP-SO400M, to ensure PRISM's robustness to different cross-modal alignment strategies.
    \item \textbf{Generalization Analysis}: For each configuration, we conduct both a systematic anisotropy diagnosis (as detailed in Section 3.2) and an end-to-end evaluation of the performance improvements yielded by PRISM-selected data.
\end{itemize}

\begin{table}[ht!]
\centering
\resizebox{\columnwidth}{!}{%
\begin{tabular}{lcc}
\toprule
\textbf{Model Family} & \textbf{Vision Encoder} & \textbf{Language Model Backbone} \\
\midrule
LLaVA-Vicuna-7B  & CLIP ViT-L/14 (336px) & Vicuna v1.5 7B \\
LLaVA-Phi2-3B    & SigLIP-SO400M & Phi-2 2.7B \\
LLaVA-Vicuna-13B & CLIP ViT-L/14 (336px) & Vicuna v1.5 13B \\
LLaVA-Qwen-7B & SigLIP-SO400M & Qwen2.5-7B (Base/Instruct) \\
LLaVA-Llama-8B  & CLIP ViT-L/14 (336px) & Llama-3-8B \\
\bottomrule
\end{tabular}%
}
\caption{\textbf{Comprehensive architectural configurations for generalization testing.} The evaluation spans multiple LLM families, varying parameter scales, and diverse vision encoders to validate the universal applicability of PRISM.}
\label{tab:model_arch}
\end{table}
\section{Impact of Textual Representation and Joint Multimodal Selection}
\label{sec:appendix_textual_analysis}

To address potential concerns regarding the role of textual prompts in data redundancy, this section provides: (1) a comparative analysis of visual vs. textual anisotropy, (2) the motivation for our visual-centric approach, and (3) additional experimental results explicitly incorporating textual features during selection.

\subsection{Contrastive Anisotropy: Visual vs. Textual Modalities}
Our diagnosis of representation anisotropy reveals a significant disparity between modalities. As illustrated in Figure 2, while visual features across all instruction datasets exhibit a non-zero mean vector $\mu$ with substantial magnitude, text features produced by the same MLLM are comparatively well-centered around the origin. Furthermore, the singular value spectrum of the textual embedding space is considerably flatter than that of the visual space. 

Because these two modalities exhibit fundamentally different geometric behaviors, applying a single global re-centering operation to concatenated multimodal features would be suboptimal. A joint mean vector would be dominated by the severe visual drift, potentially over-correcting or distorting the near-isotropic textual geometry, thereby degrading the quality of similarity estimates. Consequently, PRISM focuses its geometric correction on the visual stream, where anisotropy is the primary obstacle to redundancy detection.

\subsection{Primacy of Visual Redundancy in Instruction Tuning}
In the context of visual instruction tuning, the image typically serves as the primary carrier of new scene-level semantics. In datasets such as LLaVA-665K, text prompts often function as task-specific templates (e.g., questions or instructions) conditioned on the visual content. Therefore, redundancy in these datasets is largely governed by the visual channel: models frequently encounter near-identical visual semantics under slight variations of textual phrasing. By targeting visual anisotropy, PRISM effectively prunes redundant visual scenes that provide diminishing returns during the fine-tuning process.

\subsection{Additional Experiments with Multimodal Features}
To empirically verify this design choice, we conducted additional experiments on LLaVA-1.5-7B using features that explicitly incorporate both visual and textual information. We constructed multimodal representations by aggregating image tokens and text prompt tokens before applying the PRISM selection pipeline. 

As shown in Table~\ref{tab:multimodal_selection}, including textual prompts in the selection signal does not yield performance gains over our visual-only approach. In fact, the multimodal-feature selection (97.8\% relative performance) underperforms the visual-only PRISM (101.7\%). We hypothesize this is because text prompts introduce additional variance related to syntactic style and template length that is uncorrelated with the underlying semantic redundancy of the task. These results reinforce that visual anisotropy is the dominant factor to address for training-free intrinsic selection.

\begin{figure}[H]
    \centering
    \includegraphics[width=1\linewidth]{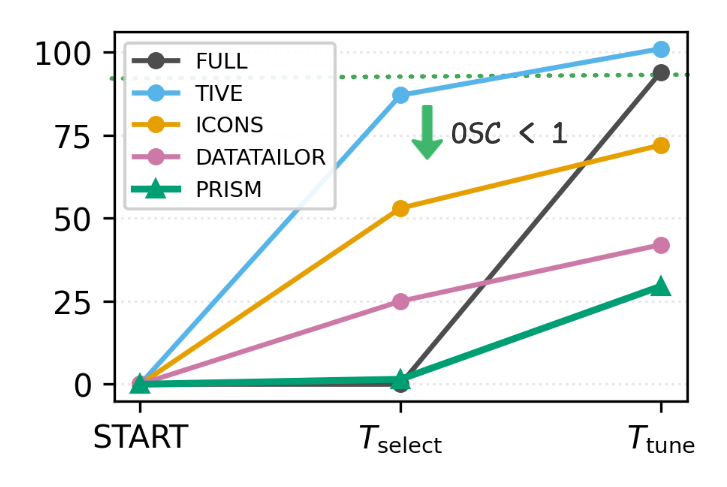}
    \caption{
    End-to-end computational cost (GPU Hours) vs. final model performance. PRISM achieves state-of-the-art performance while significantly reducing total time, thus satisfying the condition OSC $< 1$.
    }
    \label{fig:osc1}
\end{figure}

\begin{table*}[t!]
\centering
\resizebox{\textwidth}{!}{ % 调节表格宽度为文本宽度，高度按比例缩放
\begin{tabular}{l*{11}{c}}
\toprule
\textbf{Model} & \textbf{SQA} & \textbf{SQA-I} & \textbf{VizWiz} & \textbf{POPE-P} & \textbf{POPE-R} & \textbf{POPE-A} & \textbf{MM-Vet} & \textbf{MMBench} & \textbf{MME} & \textbf{MMMU} & \textbf{\cellcolor{gray!10}Rel. (\%)} \\
\midrule
LLaVA-Phi2-3B & 75.3&72.7 & 41.2&87.3 & 88.6& 86.1 &35.6 &68.7 &1765.7 &37.7 & 100\%\cellcolor{gray!10} \\

\includegraphics[height=1em]{figs/Prismatic_Star.png}\cellcolor{cyan!10}PRISM-3B & \cellcolor{cyan!10} \textbf{76.3}& \cellcolor{cyan!10} \textbf{72.8}& \cellcolor{cyan!10} 40.9 & \cellcolor{cyan!10} \textbf{87.5}& \cellcolor{cyan!10} \textbf{88.8}& \cellcolor{cyan!10} \textbf{86.5}& \cellcolor{cyan!10} 34.1& \cellcolor{cyan!10}\textbf{68.9} & \cellcolor{cyan!10} \textbf{1790.5}& \cellcolor{cyan!10} 37.6& \cellcolor{cyan!10} \textbf{100.1\%}\\
\cmidrule(lr){1-12} % 在 Full-Finetune 下面加一条分割线

LLaVA-Vicuna-7B & 69.4 & 66.8& 50.0 & 86.1& 87.3&84.2 &31.1 &64.3 &1822.6 & 35.4& 100\% \cellcolor{gray!10} \\

\includegraphics[height=1em]{figs/Prismatic_Star.png}\cellcolor{cyan!10}PRISM-7B & \textbf{71.3}\cellcolor{cyan!10} &\textbf{69.1} \cellcolor{cyan!10}  & \textbf{50.1}\cellcolor{cyan!10} &\textbf{87.7} \cellcolor{cyan!10} &\textbf{88.7} \cellcolor{cyan!10} &\textbf{85.5} \cellcolor{cyan!10} & \textbf{32.0}\cellcolor{cyan!10} &\textbf{65.2} \cellcolor{cyan!10} & \textbf{1800.0}\cellcolor{cyan!10} & 34.7\cellcolor{cyan!10} & \textbf{101.7\%}\cellcolor{cyan!10} \\
\cmidrule(lr){1-12} % 在 Full-Finetune 下面加一条分割线

LLaVA-Vicuna-13B & 74.4&71.6 &53.6 &87.4 & 88.0& 85.6&36.1  & 67.7&1826.7 & 35.1& 100\%\cellcolor{gray!10} \\

\includegraphics[height=1em]{figs/Prismatic_Star.png}\cellcolor{cyan!10}PRISM-13B  & \textbf{74.5}\cellcolor{cyan!10} &\textbf{71.8} \cellcolor{cyan!10}  & 53.1\cellcolor{cyan!10} &\textbf{87.7} \cellcolor{cyan!10} &\textbf{88.4} \cellcolor{cyan!10} &\textbf{85.7} \cellcolor{cyan!10} & \textbf{36.4}\cellcolor{cyan!10} &65.8 \cellcolor{cyan!10} & \textbf{1846.0}\cellcolor{cyan!10} & \textbf{35.7}\cellcolor{cyan!10} & \textbf{100.4\%}\cellcolor{cyan!10} \\
\cmidrule(lr){1-12} % 在 Full-Finetune 下面加一条分割线

LLaVA-Qwen-7B (Base) & 79.1 & 76.7 & 55.3 & 87.3 & 89.1 & 86.4 & 33.1 & 70.1 & 1923.1 & 46.0 & 100\%\cellcolor{gray!10} \\

\includegraphics[height=1em]{figs/Prismatic_Star.png}\cellcolor{cyan!10}PRISM-Qwen-7B (Base) & \cellcolor{cyan!10}\textbf{81.2} & \cellcolor{cyan!10}\textbf{78.9} & \cellcolor{cyan!10}55.1 & \cellcolor{cyan!10}\textbf{88.6} & \cellcolor{cyan!10}\textbf{90.3} & \cellcolor{cyan!10}\textbf{87.5} & \cellcolor{cyan!10}\textbf{33.7} & \cellcolor{cyan!10}\textbf{71.5} & \cellcolor{cyan!10}\textbf{1895.8} & \cellcolor{cyan!10}45.2 & \cellcolor{cyan!10}\textbf{101.0\%} \\
\cmidrule(lr){1-12} % 在 Full-Finetune 下面加一条分割线

LLaVA-Qwen-7B (Instruct) & 80.2 & 77.8 & 54.7 & 87.9 & 89.7 & 86.9 & 33.8 & 71.0 & 1935.4 & 46.8 & 100\%\cellcolor{gray!10} \\

\includegraphics[height=1em]{figs/Prismatic_Star.png}\cellcolor{cyan!10}PRISM-Qwen-7B (Instruct) & \cellcolor{cyan!10}\textbf{82.3} & \cellcolor{cyan!10}\textbf{80.1} & \cellcolor{cyan!10}54.2 & \cellcolor{cyan!10}\textbf{89.1} & \cellcolor{cyan!10}\textbf{90.8} & \cellcolor{cyan!10}\textbf{88.1} & \cellcolor{cyan!10}\textbf{34.2} & \cellcolor{cyan!10}\textbf{72.4} & \cellcolor{cyan!10}\textbf{1908.6} & \cellcolor{cyan!10}46.1 & \cellcolor{cyan!10}\textbf{100.9\%} \\
\cmidrule(lr){1-12} % 在 Full-Finetune 下面加一条分割线

LLaVA-Llama-8B & 77.6 & 75.2 & 53.7 & 86.3 & 87.6 & 84.0 & 31.8 & 68.5 & 1889.7 & 40.3 & 100\%\cellcolor{gray!10} \\

\includegraphics[height=1em]{figs/Prismatic_Star.png}\cellcolor{cyan!10}PRISM-Llama-8B & \cellcolor{cyan!10}\textbf{79.4} & \cellcolor{cyan!10}\textbf{77.3} & \cellcolor{cyan!10}53.4 & \cellcolor{cyan!10}\textbf{87.8} & \cellcolor{cyan!10}\textbf{89.1} & \cellcolor{cyan!10}\textbf{85.6} & \cellcolor{cyan!10}31.5 & \cellcolor{cyan!10}\textbf{69.9} & \cellcolor{cyan!10}\textbf{1862.5} & \cellcolor{cyan!10}39.7 & \cellcolor{cyan!10}\textbf{100.8\%} \\

\bottomrule
\end{tabular}
}
\caption{
\textbf{Cross-Model Generalization and Scalability of PRISM.}
PRISM's selection mechanism generalizes robustly across diverse MLLM architectures by correcting representation anisotropy, as shown across five LLaVA-style model--encoder combinations.
By identifying semantically dense samples for each architecture, PRISM consistently improves performance over full finetuning for both small and large language models—including Vicuna, Phi-2, Qwen2.5, and Llama-3—as well as distinct vision encoders such as CLIP and SigLIP.
Together with Fig.~\ref{fig:archnew}, these results confirm that PRISM’s gains are not tied to a particular LLaVA configuration but transfer across diverse multimodal model families.
}
\label{tab:arch}
\end{table*}
\begin{table*}[ht]
\centering
\resizebox{\textwidth}{!}{ 
\begin{tabular}{l*{12}{c}}
\toprule
\textbf{Method} & \textbf{SQA} & \textbf{SQA-I} & \textbf{VizWiz} & \textbf{POPE-P} & \textbf{POPE-R} & \textbf{POPE-A} & \textbf{MM-Vet} & \textbf{MMBench} & \textbf{MME-P} & \textbf{MME-C} & \textbf{MMMU} & \textbf{Rel. (\%)} \\
\midrule
Full-Finetune & 69.4 & 66.8 & 50.0 & 86.1 & 87.3 & 84.2 & 31.1 & 64.3 & 1510.7 & 311.9 & 35.4 & 100\% \\
Multimodal PRISM & 68.2 & 65.1 & 48.9 & 85.3 & 86.0 & 83.1 & 30.2 & 63.0 & 1455.0 & 305.0 & 34.1 & 97.8\% \\
\cmidrule(lr){1-13}
\raisebox{-0.2em}{\includegraphics[height=0.8em]{figs/12483567.png}} \textbf{Visual-only PRISM} & \textbf{71.3}\cellcolor{cyan!10} & \textbf{69.1}\cellcolor{cyan!10} & \textbf{50.1}\cellcolor{cyan!10} & \textbf{87.7}\cellcolor{cyan!10} & \textbf{88.7}\cellcolor{cyan!10} & \textbf{85.5}\cellcolor{cyan!10} & \textbf{32.0}\cellcolor{cyan!10} & \textbf{65.2}\cellcolor{cyan!10} & 1470.0\cellcolor{cyan!10} & \textbf{330.0}\cellcolor{cyan!10} & 34.7\cellcolor{cyan!10} & \textbf{101.7\%}\cellcolor{cyan!10} \\
\bottomrule
\end{tabular}
}
\caption{
    \textbf{Comparison of selection using visual-only features versus joint multimodal features.} The visual-centric approach (PRISM) demonstrates higher stability as text prompts often introduce non-semantic noise.
}
\label{tab:multimodal_selection}
\end{table*}
\section{Limitations of Existing Methods and Our Geometric Motivation}
\label{sec:limitations_and_motivation}

This section provides a detailed analysis of the structural limitations inherent in existing visual instruction selection paradigms and introduces the theoretical motivation for our work. We first outline the three primary failure modes of prior methods and then present our diagnosis that these issues stem from a single, unaddressed geometric flaw in MLLM feature representations.

\subsection{A Critical Review of Existing Selection Paradigms}
Despite their diverse approaches, current selection methods consistently exhibit three critical drawbacks that limit their practical utility, often violating the core principles of effective data selection we established in Section~\ref{principle}.

\noindent\textbf{1. Performance Degradation.}
A primary goal of data selection is to curate a smaller, more potent dataset that ideally improves model performance. However, as shown in our main results (Table~\ref{tab:main}), many existing methods paradoxically lead to models that underperform a full fine-tuning baseline. This outcome directly contradicts our \textit{Principle 1 (Performance Fidelity)}, suggesting that their heuristics for identifying "valuable" data are often misaligned with the true learning objectives.

\noindent\textbf{2. Prohibitive Computational Cost.}
The second major failure mode is excessive computational overhead, particularly in Training-Based methods. For instance, techniques like TIVE~\citep{liu2024morehighvaluedataselection} require a costly "warm-up" phase, often involving LoRA-based~\citep{hu2021lora,bi-etal-2025-llava} fine-tuning, simply to generate the gradient signals needed for selection. As we show in our runtime analysis (Table~\ref{tab:time}), the time spent on selection alone can exceed the time saved during the subsequent tuning, violating \textit{Principle 2 (Net Efficiency Gain)} and rendering the approach impractical at scale.

\noindent\textbf{3. Proxy Model Bias.}
To avoid the high costs of gradient computation, some methods delegate scoring to an external proxy model (e.g., GPT-4 or a reward model). While faster, this introduces a significant bias, as the proxy's scoring function may not align with the target MLLM's specific learning needs. This approach yields a generic, one-size-fits-all dataset, failing to provide an optimal, tailored subset for the specific MLLM being trained.

\subsection{The Unifying Cause: A Deep Dive into Representation Anisotropy}
We argue that these seemingly disparate issues are symptoms of a single, deeper problem: a failure to account for the pathological geometry of MLLM feature spaces. In practice, high-dimensional embeddings learned by deep networks are rarely isotropic (uniformly distributed); instead, they are highly \textbf{anisotropic}, exhibiting a strong directional bias that corrupts the very notion of semantic distance and plagues models across modalities~\citep{ethayarajh-2019-contextual, jing2021understanding}.

\paragraph{The Geometry of Anisotropy.} An embedding space is said to be \textbf{isotropic} if its features are, on average, uniformly distributed in all directions from the origin. For a zero-mean feature set $\mathcal{X} = \{\mathbf{x}_i\}$, this property is formally defined by a covariance matrix proportional to the identity matrix, $\mathbb{E}[\mathbf{x}\mathbf{x}^\top] \propto \mathbf{I}$. Such a space leverages its full dimensionality to represent concepts. In stark contrast, an \textbf{anisotropic} space is characterized by a non-uniform, biased distribution. 

\noindent\textbf{Low Effective Rank and Representation Degeneration.} A complementary perspective comes from singular value decomposition (SVD). In an anisotropic space, the singular value spectrum of the centered feature matrix exhibits a sharp, steep decay. As illustrated in Figure~\ref{fig:miu}, a few dominant singular values capture most of the data's variance, while the rest are negligible. This indicates a \textbf{low effective rank}~\citep{arora-etal-2016-latent}, meaning the representation fails to utilize the full capacity of its embedding dimensions. This phenomenon, often called \textbf{representation degeneration}, is prevalent in modern architectures, including those trained with contrastive objectives like CLIP~\citep{radford2021learning}.

\paragraph{Implications for Data Selection.} This unaddressed geometric flaw helps explain the practical limitations of prior art. Because simple and efficient geometric metrics (e.g., cosine similarity, farthest-point sampling) become unreliable under strong anisotropy, many existing methods turn to more expensive proxy signals to approximate semantic value. Training-Based methods pay a heavy price in computation to derive signals from the model's dynamics, while Proxy-Based methods introduce potential biases from an external scorer. Both are, in effect, paying a high computational "tax" for not first correcting the geometry of the space itself. Our work is motivated by the insight that by directly diagnosing and correcting this anisotropy, we can unlock a far more efficient and principled path to data selection that operates on the intrinsic, geometrically-sound structure of the data.
\section{Theoretical Foundation: Anisotropy as a Source of Computational Overhead}
\label{sec:appendix_theory}

This section analyzes how representation anisotropy can make existing data selection methods for MLLMs inefficient in practice. We formalize the concept of representation anisotropy and show how it degrades simple geometric metrics for measuring semantic diversity. We then discuss why, in the absence of geometric correction, many prior methods rely on proxy signals or training dynamics that introduce substantial overhead, motivating our geometry-aware approach.

\subsection{The Geometric Pathology of Anisotropy}
We begin by formalizing the geometric properties of the feature space. Let $\phi: \mathcal{D} \to \mathbb{R}^d$ be the MLLM's embedding function. The observed anisotropy in the visual feature space implies a non-zero mean vector $\boldsymbol{\mu} = \mathbb{E}[\phi(d)]$ with a large magnitude. Consequently, any feature vector $\mathbf{x}_i = \phi(d_i)$ can be decomposed into a shared, dominant global component and a sample-specific semantic residual:
\begin{equation}
    \mathbf{x}_i = \boldsymbol{\mu} + \boldsymbol{\delta}_i
\end{equation}
where $\boldsymbol{\delta}_i$ encodes the unique semantics of sample $d_i$. The defining characteristic of a highly anisotropic space is that this shared component dominates the representation, i.e., $\|\boldsymbol{\mu}\|_2 \gg \mathbb{E}[\|\boldsymbol{\delta}_i\|_2]$. This property has a catastrophic consequence for standard geometric metrics, which we formalize in the following theorem.

\begin{theorem}[Corruption of Geometric Proximity]
\label{thm:contamination_appendix}
In an anisotropic representation space where $\|\boldsymbol{\mu}\|_2 \gg \mathbb{E}[\|\boldsymbol{\delta}_i\|_2]$, the cosine similarity between any two randomly sampled vectors $\mathbf{x}_i$ and $\mathbf{x}_j$ is dominated by the global drift vector $\boldsymbol{\mu}$, masking their true semantic dissimilarity, which is encoded in the residuals $\boldsymbol{\delta}_i, \boldsymbol{\delta}_j$.
\end{theorem}
\begin{proof}
The cosine similarity is given by:
\begin{equation}
\cos(\mathbf{x}_i, \mathbf{x}_j) = \frac{(\boldsymbol{\mu} + \boldsymbol{\delta}_i)^\top (\boldsymbol{\mu} + \boldsymbol{\delta}_j)}{\|\boldsymbol{\mu} + \boldsymbol{\delta}_i\|_2 \|\boldsymbol{\mu} + \boldsymbol{\delta}_j\|_2}
\end{equation}
The numerator expands to $\|\boldsymbol{\mu}\|_2^2 + \boldsymbol{\mu}^\top(\boldsymbol{\delta}_i + \boldsymbol{\delta}_j) + \boldsymbol{\delta}_i^\top\boldsymbol{\delta}_j$. Since $\|\boldsymbol{\mu}\|_2$ is the dominant term, the denominator can be approximated using a first-order Taylor expansion as $\|\boldsymbol{\mu} + \boldsymbol{\delta}_i\|_2 \approx \|\boldsymbol{\mu}\|_2 + \frac{\boldsymbol{\mu}^\top\boldsymbol{\delta}_i}{\|\boldsymbol{\mu}\|_2}$. Substituting these approximations and retaining only the leading-order terms yields:
\begin{align}
\cos(\mathbf{x}_i, \mathbf{x}_j)
&\approx 
\frac{\|\boldsymbol{\mu}\|_2^2}{
\left(\|\boldsymbol{\mu}\|_2 + \frac{\boldsymbol{\mu}^\top\boldsymbol{\delta}_i}{\|\boldsymbol{\mu}\|_2}\right)
\left(\|\boldsymbol{\mu}\|_2 + \frac{\boldsymbol{\mu}^\top\boldsymbol{\delta}_j}{\|\boldsymbol{\mu}\|_2}\right)} \nonumber\\
&\approx 
1 - \frac{1}{2}\left\|
\frac{\boldsymbol{\delta}_i'}{\|\boldsymbol{\mu}\|_2}
- 
\frac{\boldsymbol{\delta}_j'}{\|\boldsymbol{\mu}\|_2}
\right\|_2^2 
\approx 1
\end{align}

where $\boldsymbol{\delta}'$ is the component of $\boldsymbol{\delta}$ orthogonal to $\boldsymbol{\mu}$. The true semantic similarity, proportional to $\boldsymbol{\delta}_i^\top\boldsymbol{\delta}_j$, is relegated to lower-order terms that are numerically insignificant, rendering the metric incapable of distinguishing unique samples.
\end{proof}

\subsection{Practical Cost of Ignoring Anisotropy}
The failure of simple geometric proxies, as established in Theorem~\ref{thm:contamination_appendix}, creates a practical dilemma for data selection. When raw feature distances are dominated by global drift, reliable semantic selection typically requires either correcting the geometry or introducing auxiliary signals to approximate the latent semantic structure of the data. This practical cost account explains why many existing visual-instruction selection pipelines incur substantial overhead.

Common practical strategies include:
\begin{enumerate}
    \item \textbf{Implicit Structure Recovery:} Employing iterative clustering or concept-discovery algorithms to recover semantic groups in a distorted space. Such procedures typically introduce costs such as $\mathcal{O}(I \cdot N \cdot K \cdot d)$, where $I$ is the number of iterations and $K$ is the number of clusters.
    \item \textbf{Bypassing Geometry via Training Dynamics:} Using model gradients or training loss as a proxy for data value. These signals can be effective but require forward and/or backward passes over the data, with costs on the order of $\mathcal{O}(N \cdot C_{\text{pass}})$, where $C_{\text{pass}}$ is the cost of a model pass.
\end{enumerate}

Methods such as GraNd~\citep{paul2023deeplearningdatadiet} and TIVE~\citep{liu2024morehighvaluedataselection} exemplify the second pattern: they obtain useful value estimates from training dynamics, but the resulting selection stage can become comparable to, or even exceed, the cost saved by later training on a smaller subset. PRISM takes a different route. By removing the dominant first-order drift before scoring redundancy, it corrects the usefulness of inexpensive correlation-based geometry and avoids these additional proxy or training-dynamics costs.
\section{Cross-Architecture Diagnosis of Representation Anisotropy}
\label{sec:appendix_anisotropy_stats}

To further validate that representation anisotropy is a universal characteristic of Multimodal Large Language Models (MLLMs), we expand our diagnostic analysis across five distinct model-encoder combinations. This section provides detailed statistical distributions of per-dimension means to characterize the geometric shift in the feature space.

\subsection{Metric Definitions}
We utilize the following statistics to quantify the distribution of per-dimension means across different datasets:
\begin{itemize}
    \item \textbf{Min / Max}: The extreme values of the per-dimension mean vectors.
    \item \textbf{Mean($|x|$)}: The arithmetic mean of the absolute per-dimension means, serving as a primary indicator of the magnitude of the global shift from the origin.
    \item \textbf{Percentiles (P25, P75, P99)}: Values representing the spread and skewness of the mean distribution across dimensions.
\end{itemize}

\subsection{Per-dimension Mean Statistics across Configurations}
Tables \ref{tab:stats_vicuna} through \ref{tab:stats_llama} present the statistics for diverse language backbones and vision encoders. Across all configurations, a consistent pattern emerges: textual features remain well-centered around zero ($\text{Mean}(|x|) \approx 0.11\text{--}0.13$), whereas all visual instruction datasets exhibit substantial non-zero shifts ($\text{Mean}(|x|) \approx 0.36\text{--}0.66$).
In an ideal isotropic space, such as the space induced by high-quality text embeddings, the global mean vector should be negligible ($\|\boldsymbol{\mu}\|_2 \approx 0$), and the representation should be dominated by sample-specific variance components $\boldsymbol{\delta}_i$. However, current visual representations from CLIP- and SigLIP-based encoders deviate significantly from this ideal. The per-dimension mean statistics reported here, together with the singular value distributions in Figure~\ref{fig:miu}, show a consistent and substantial non-zero mean shift in visual features. This empirical dominance of the global component directly motivates the condition $\|\boldsymbol{\mu}\|_2 \gg \mathbb{E}[\|\boldsymbol{\delta}_i\|_2]$ in Theorem~\ref{thm:contamination}, modeling the observed geometry of current visual encoders rather than an arbitrary theoretical convenience.

\subsection{Dataset-Mean Centering Sanity Check}
To verify that the observed shift is not merely a small non-zero mean, we further analyze OCRVQA layer-1 features with $N=500$ samples and $D=4096$ dimensions. The raw corpus mean has norm $\|\boldsymbol{\mu}\|_2=50.8456$, while dataset-mean centering reduces it to $1.8645\times 10^{-6}$. More importantly, the mean component accounts for $0.9721$ of the origin second moment:
\begin{equation}
    \frac{\|\boldsymbol{\mu}\|_2^2}{\operatorname{tr}(\mathbb{E}[\mathbf{x}\mathbf{x}^\top])}=0.9721.
\end{equation}
Since $\operatorname{tr}(\mathbb{E}[\mathbf{x}\mathbf{x}^\top])=\|\boldsymbol{\mu}\|_2^2+\mathbb{E}[\|\mathbf{x}-\boldsymbol{\mu}\|_2^2]$, this implies
\begin{equation}
    \frac{\|\boldsymbol{\mu}\|_2^2}{\mathbb{E}[\|\boldsymbol{\delta}\|_2^2]}\approx \frac{0.9721}{0.0279}\approx 34.8,
    \quad
    \frac{\|\boldsymbol{\mu}\|_2}{\sqrt{\mathbb{E}[\|\boldsymbol{\delta}\|_2^2]}}\approx 5.9.
\end{equation}
Thus, the mean term is not only non-zero but dominates the sample-specific residual energy, directly supporting the assumption $\|\boldsymbol{\mu}\|_2^2 \gg \mathbb{E}[\|\boldsymbol{\delta}_i\|_2^2]$ used in our analysis.

% Table 1: Vicuna + CLIP
\begin{table}[ht!]
\centering
\caption{\textbf{Statistics for Vicuna v1.5 7B + CLIP ViT-L/14 336px.}}
\label{tab:stats_vicuna}
\resizebox{\columnwidth}{!}{
\begin{tabular}{lcccccc}
\toprule
\textbf{Dataset} & \textbf{Min} & \textbf{P25} & \textbf{P75} & \textbf{P99} & \textbf{Max} & \textbf{Mean($|x|$)} \\
\midrule
Text & -2.8851 & -0.1033 & 0.0064 & 0.1105 & 0.3956 & 0.1310 \\
A-OKVQA & -4.0186 & -0.3351 & 0.0052 & 0.3329 & 1.2545 & 0.4045 \\
GQA & -4.0984 & -0.3448 & 0.0057 & 0.3404 & 1.2815 & 0.4141 \\
LLaVA & -4.0885 & -0.3427 & 0.0046 & 0.3394 & 1.2819 & 0.4126 \\
OCRVQA & -5.6103 & -0.5129 & 0.0023 & 0.5187 & 1.8318 & 0.6127 \\
VG & -4.1034 & -0.3449 & 0.0062 & 0.3413 & 1.2827 & 0.4147 \\
\bottomrule
\end{tabular}
}
\end{table}

% Table 2: Phi-2 + SigLIP
\begin{table}[ht!]
\centering
\caption{\textbf{Statistics for Phi-2 2.7B + SigLIP-SO400M-Patch14-384.}}
\label{tab:stats_phi2}
\resizebox{\columnwidth}{!}{
\begin{tabular}{lcccccc}
\toprule
\textbf{Dataset} & \textbf{Min} & \textbf{P25} & \textbf{P75} & \textbf{P99} & \textbf{Max} & \textbf{Mean($|x|$)} \\
\midrule
Text & -2.9124 & -0.1047 & 0.0072 & 0.1119 & 0.4012 & 0.1328 \\
A-OKVQA & -4.3215 & -0.3587 & 0.0048 & 0.3612 & 1.3421 & 0.4384 \\
GQA & -4.4102 & -0.3701 & 0.0053 & 0.3689 & 1.3745 & 0.4501 \\
LLaVA & -4.4018 & -0.3675 & 0.0042 & 0.3668 & 1.3751 & 0.4483 \\
OCRVQA & -6.0321 & -0.5514 & 0.0019 & 0.5589 & 1.9647 & 0.6589 \\
VG & -4.4147 & -0.3704 & 0.0058 & 0.3695 & 1.3754 & 0.4505 \\
\bottomrule
\end{tabular}
}
\end{table}

% Table 3: Qwen2.5 Base
\begin{table}[ht!]
\centering
\caption{\textbf{Statistics for Qwen2.5 7B Base + CLIP ViT-L/14 336px.}}
\label{tab:stats_qwen_base}
\resizebox{\columnwidth}{!}{
\begin{tabular}{lcccccc}
\toprule
\textbf{Dataset} & \textbf{Min} & \textbf{P25} & \textbf{P75} & \textbf{P99} & \textbf{Max} & \textbf{Mean($|x|$)} \\
\midrule
Text & -2.6734 & -0.0891 & 0.0058 & 0.0945 & 0.3512 & 0.1089 \\
A-OKVQA & -3.7845 & -0.3012 & 0.0047 & 0.3089 & 1.1523 & 0.3698 \\
GQA & -3.8621 & -0.3101 & 0.0052 & 0.3154 & 1.1789 & 0.3789 \\
LLaVA & -3.8534 & -0.3079 & 0.0041 & 0.3147 & 1.1793 & 0.3776 \\
OCRVQA & -5.1987 & -0.4687 & 0.0021 & 0.4823 & 1.6845 & 0.5634 \\
VG & -3.8665 & -0.3104 & 0.0056 & 0.3162 & 1.1795 & 0.3793 \\
\bottomrule
\end{tabular}
}
\end{table}

% Table 4: Llama-3
\begin{table}[ht!]
\centering
\caption{\textbf{Statistics for Llama-3-8B + CLIP ViT-L/14 336px.}}
\label{tab:stats_llama}
\resizebox{\columnwidth}{!}{
\begin{tabular}{lcccccc}
\toprule
\textbf{Dataset} & \textbf{Min} & \textbf{P25} & \textbf{P75} & \textbf{P99} & \textbf{Max} & \textbf{Mean($|x|$)} \\
\midrule
Text & -2.6512 & -0.0876 & 0.0055 & 0.0929 & 0.3445 & 0.1067 \\
A-OKVQA & -3.7621 & -0.2989 & 0.0046 & 0.3067 & 1.1434 & 0.3667 \\
GQA & -3.8389 & -0.3078 & 0.0051 & 0.3132 & 1.1701 & 0.3756 \\
LLaVA & -3.8301 & -0.3056 & 0.0040 & 0.3125 & 1.1705 & 0.3743 \\
OCRVQA & -5.1623 & -0.4634 & 0.0020 & 0.4789 & 1.6712 & 0.5589 \\
VG & -3.8432 & -0.3081 & 0.0055 & 0.3140 & 1.1707 & 0.3760 \\
\bottomrule
\end{tabular}
}
\end{table}

\subsection{Architectural Generality and Observations}
The results demonstrate that representation anisotropy is a robust and persistent phenomenon across different language backbones (Vicuna, Phi-2, Qwen, Llama) and vision encoders (CLIP, SigLIP). While more advanced language models slightly reduce the absolute magnitude of visual anisotropy (e.g., Llama-3 and Qwen show approximately 10\% lower $\text{Mean}(|x|)$ than Vicuna), the visual-textual gap remains pronounced in all cases. This evidence confirms that the geometric origins of Global Semantic Drift are general to the MLLM architecture and are not specific to the LLaVA framework.

\section{Benchmark Details}
\label{sec:benchmark_details}

We provide detailed descriptions of the seven multimodal benchmarks used in our main evaluation.

\noindent\textbf{MMBench}~\citep{liu2024mmbenchmultimodalmodelallaround} is a multi-faceted benchmark that evaluates the all-around capabilities of MLLMs. It assesses both perception and reasoning through single-choice questions organized hierarchically across a wide range of domains.

\noindent\textbf{ScienceQA}~\citep{lu2022learn} evaluates scientific question answering. It features multimodal questions from various science domains that require models to jointly reason over visual and textual contexts, often involving diagrams, charts, and complex text.

\noindent\textbf{MME}~\citep{yin2023survey} is a comprehensive benchmark designed to assess multimodal reasoning and perception skills. It comprises 14 sub-tasks that cover a broad spectrum of capabilities, from object existence and position to text translation and code reasoning.

\noindent\textbf{POPE}~\citep{li2023evaluatingobjecthallucinationlarge} is an evaluation framework specifically designed to measure object hallucination in MLLMs. It uses a polling-based (Yes/No) question-answering approach to probe whether a model falsely perceives objects that are not present in an image.

\noindent\textbf{VizWiz}~\citep{gurari2018vizwizgrandchallengeanswering} tests a model's ability to answer visual questions taken "in-the-wild" by visually impaired users. It serves as a challenging test for zero-shot generalization, as the questions and images are often unconventional and noisy.

\noindent\textbf{MM-Vet}~\citep{yu2024mmvetevaluatinglargemultimodal} is an evaluation benchmark designed to measure the emergent, instruction-following capabilities of MLLMs in a multimodal context. It assesses a model's ability to handle complex prompts that integrate perception, reasoning, and world knowledge.

\noindent\textbf{MMMU}~\citep{yue2024mmmumassivemultidisciplinemultimodal} is a massive, multi-discipline benchmark that assesses expert-level MLLM capabilities. It requires the integration of perception, knowledge, and reasoning across a wide array of college-level subjects, from science and engineering to arts and humanities.

\section{Sensitivity Analysis of the Selection Ratio}
\label{sec:appendix_tau_analysis}

In this section, we provide a detailed sensitivity analysis of the selection ratio $\tau$, which governs the proportion of data retained for instruction tuning. Unlike static hyperparameters, $\tau$ serves as a flexible selection budget that practitioners can adjust based on specific computational constraints.
Table~\ref{tab:tau_robustness} presents the performance trade-offs across various sampling ratios. Our analysis reveals that PRISM is robust across diverse selection budgets:
\begin{itemize}
    \item \textbf{Monotonic Visual Scaling}: Multimodal performance scales consistently with the data budget, with the 30\% ratio (PRISM-Instruct-250K) effectively surpassing the full-dataset baseline.
    \item \textbf{Language Capability Preservation}: Textual performance remains remarkably high even at extreme pruning levels (e.g., 5\%), suggesting that PRISM's redundancy identification excels at isolating a high-quality "data core" that minimizes catastrophic forgetting.
\end{itemize}

\begin{table}[t!]
\centering
\resizebox{\columnwidth}{!}{ 
\begin{tabular}{l*{3}{c}}
\toprule
\textbf{Sampling Ratio ($\tau$)} & \textbf{Visual Performance} & \textbf{Textual Performance} \\
\midrule
5\% & 92.3 & 104.4 \\
10\% & 95.1 & 102.9 \\
20\% & 97.1 & 102.1 \\
\cmidrule(lr){1-3}
\raisebox{-0.2em}{\includegraphics[height=0.8em]{figs/12483567.png}} \textbf{30\% (PRISM Default)} & \textbf{101.7}\cellcolor{cyan!10} & \textbf{101.9}\cellcolor{cyan!10} \\
\cmidrule(lr){1-3}
Full Fine-tuning (Baseline) & 100.0 & 100.0 \\
\bottomrule
\end{tabular}
}
\caption{
    \textbf{Performance robustness across varied selection ratios $\tau$.} Visual performance denotes the relative average score across multimodal benchmarks, while textual performance reflects foundational language understanding scores relative to the baseline.
}
\label{tab:tau_robustness}
\end{table}

\section{Selection Metric Rationale: Correlation vs. Second-Order Distance}
\label{sec:appendix_metric_rationale}

The design of PRISM prioritizes a training-free, computationally efficient pipeline. While second-order measures like Mahalanobis distance are theoretically relevant to anisotropy, they introduce significant bottlenecks in large-scale multimodal settings. We justify the use of implicit re-centering via Pearson-style correlation based on the following factors:

\subsection{Algorithmic Efficiency and Scalability}
The estimation of Mahalanobis distance requires the computation and inversion of a covariance matrix $\Sigma \in \mathbb{R}^{d \times d}$. For modern MLLMs with high-dimensional feature spaces, this incurs an $\mathcal{O}(Nd^2)$ cost for estimation and $\mathcal{O}(d^3)$ for inversion. In contrast, PRISM uses corpus-mean re-centering and an exact aggregate form of Pearson-style normalized inner-product scoring, giving an $\mathcal{O}(Nd)$ scoring cost and enabling the processing of the LLaVA-665K dataset in just 1.5 GPU hours.

\subsection{Numerical Stability in Anisotropic Spaces}
Our singular value analysis demonstrates that pre-trained visual features occupy a low effective rank, where variance is concentrated in a few dominant dimensions. This geometric degeneration often results in ill-conditioned covariance matrices. Relying on matrix inversion in such spaces would require additional regularization (e.g., eigenvalue clipping or shrinkage), introducing new hyperparameters and compromising the simplicity of the training-free paradigm.

\subsection{Complexity of Correlation Computation}

Although the redundancy score is written as an average over correlations, PRISM evaluates it through an exact aggregate form rather than enumerating all sample pairs. After global re-centering and normalization, each correlation is an inner product between transformed features, so the average correlation of every sample can be obtained from a single global normalized-feature mean. This yields an $\mathcal{O}(Nd)$ scoring procedure and avoids materializing any $N \times N$ matrix. The full derivation, numerical-stability details, and chunked GPU implementation are provided in Appendix~\ref{sec:runtime_details}.

\section{Qualitative Analysis}
\label{sec:appendix_qualitative}

To provide a more intuitive understanding of the Intrinsic Selection paradigm, we conduct a qualitative analysis of samples ranked by PRISM. Figure~\ref{fig:case_study} illustrates representative examples from the top-ranked (selected) and bottom-ranked (pruned) tiers of the LLaVA-665K dataset.

\subsection{High-Value: Challenging and Informative}
The top-ranked samples prioritized by PRISM typically involve \textbf{fine-grained perception} and \textbf{complex contextual reasoning}. For instance, in the "Asia street" example (Figure~\ref{fig:case_study}, Left), the model must identify localized visual cues such as the road marking Tomare, the specific structure of utility poles, and architectural styles to correctly determine the geographical region. 
 From a training perspective, these informative interactions provide a stronger learning signal, often corresponding to higher initial loss values, which is critical for mastering expert-level reasoning.

\subsection{Low-Value: Trivial and Redundant}
Conversely, bottom-ranked samples are frequently characterized by \textbf{trivial perception} or \textbf{high semantic redundancy}. In the "Sunny weather" example (Figure~\ref{fig:case_study}, Right), the task involves identifying a generic weather condition in a common park setting. Such visual scenes—comprising grass, trees, and sunlight—are over-represented in the corpus, leading to a high correlation with the Global Semantic Drift.

Because these samples offer saturated information and redundant gradients, training on them yields diminishing returns and may even exacerbate "knowledge forgetting". By pruning these trivial samples, PRISM ensures a more efficient use of the computational budget without sacrificing model performance.

\section{Detailed Runtime Analysis}
\label{sec:runtime_details}

\subsection{Wall-Clock Runtime Breakdown}

To provide a transparent account of the computational costs associated with different selection methods, we present a detailed breakdown of the wall-clock runtime in Table~\ref{tab:time}. All experiments were conducted on an NVIDIA A100 80GB GPU, and time was measured in hours.

The table decomposes the total pipeline duration into two key phases: \textbf{Data Selection} (the cost of the selection method itself) and \textbf{Visual Instruction Tuning} (the cost of fine-tuning the model on the selected data subset).

\begin{itemize}
    \item \textbf{Full-Finetune:} This serves as our baseline, requiring 94 GPU hours for tuning on the complete dataset, with zero selection cost.

    \item \textbf{TIVE:} This Training-Based method exemplifies the high overhead of existing approaches. The selection phase alone consumes 87 hours, which is 92\% of the baseline's entire tuning time. Although the final tuning is fast (14 hours), the total pipeline (101 hours) is 7.5\% slower than simply fine-tuning on all data, resulting in an OSC score greater than 1.

    \item \textbf{PRISM:} Our method demonstrates exceptional efficiency. The entire selection process, including feature extraction and correlation computation, takes only 1.5 hours. This allows for a total pipeline time of just 29.5 hours, a \textbf{71\% reduction} compared to the full fine-tuning baseline, robustly satisfying our efficiency principles.
\end{itemize}

\begin{table}[ht]
\centering
\resizebox{\columnwidth}{!}{
\begin{tabular}{l*{3}{c}}
\toprule
\textbf{Method} & \textbf{Data Selection} & \textbf{Visual Instruction Tuning} & \textbf{Overall} \\
\midrule
Full-Finetune & - & 94 (Hours)& 94 \\
\cmidrule(lr){1-4}

TIVE & 87 (Hours) & 14 (Hours) & 101 (\textcolor{red}{+7.5\%})\\

\includegraphics[height=1em]{figs/Prismatic_Star.png}PRISM & 1.5 (Hours) & 28 (Hours)& 29.5 (\textcolor{green}{-71\%})\\
\bottomrule
\end{tabular}
}
\caption{Wall-clock runtime breakdown on an NVIDIA A100 80GB GPU. All times are measured in hours. Percentages in the 'Overall' column indicate the change relative to the Full-Finetune baseline.}
\label{tab:time}
\end{table}

\subsection{Pairwise Redundancy Objective and Linear-Time Computation}

A natural concern is that the redundancy score in Eq.~\ref{eq:redundancy_score} appears to require all pairwise correlations. PRISM's conceptual objective is the full pairwise redundancy objective over the re-centered corpus. Let $K_{ij}=\rho(\mathbf{F}_i,\mathbf{F}_j)$ denote the centered cosine affinity between two samples. Then $\mathcal{R}(d_i)=\frac{1}{N-1}\sum_{j\neq i}K_{ij}$ is the normalized weighted degree of $d_i$ in the complete corpus-level similarity graph, equivalently a kernel-density estimate in the re-centered feature space. A high value means that the sample lies in a dense, repeatedly covered semantic region; a low value means that it has weak affinity to the dominant redundant modes and therefore increases the selected subset's coverage of less duplicated visual semantics. This is why selecting the low-correlation tail directly targets novelty, diversity, and reduced over-representation in the instruction pool.

The aggregate expression below provides an algebraic acceleration of this complete-graph score. Define the centered and normalized feature
\begin{equation}
    \mathbf{g}_i =
    \frac{\mathbf{F}_i - \boldsymbol{\mu}_{\mathcal{F}}}
    {\|\mathbf{F}_i - \boldsymbol{\mu}_{\mathcal{F}}\|_2},
\end{equation}
so that $\rho(\mathbf{F}_i,\mathbf{F}_j)=\mathbf{g}_i^\top\mathbf{g}_j$. Let $\bar{\mathbf{g}}=\frac{1}{N}\sum_{j=1}^{N}\mathbf{g}_j$ be the sufficient statistic that sums all normalized centered features. Then the same leave-one-out pairwise average can be evaluated as
\begin{align}
    \mathcal{R}(d_i)
    &= \frac{1}{N-1}\sum_{j\neq i}\mathbf{g}_i^\top\mathbf{g}_j \nonumber\\
    &= \frac{\mathbf{g}_i^\top\sum_{j=1}^{N}\mathbf{g}_j-\mathbf{g}_i^\top\mathbf{g}_i}{N-1} \nonumber\\
    &= \frac{N\mathbf{g}_i^\top\bar{\mathbf{g}}-1}{N-1},
\end{align}
where the last equality follows from $\|\mathbf{g}_i\|_2=1$. The vector $\bar{\mathbf{g}}$ is a sufficient statistic that compresses the complete pairwise graph under the linear centered-cosine kernel. It preserves the exact ranking induced by the full average-correlation score while avoiding explicit enumeration of all pairs. Once $\boldsymbol{\mu}_{\mathcal{F}}$ and this sufficient statistic are computed, the $N$ complete-graph degree scores can be obtained in linear time.

In implementation, PRISM processes the feature matrix in chunks. The first streaming pass accumulates the corpus-level mean $\boldsymbol{\mu}_{\mathcal{F}}$. The second pass centers and normalizes each chunk, accumulates $\bar{\mathbf{g}}$, and optionally writes the normalized chunk to disk. The final pass streams stored or recomputed normalized chunks, computes $\mathbf{g}_i^\top\bar{\mathbf{g}}$ for each sample, and writes the redundancy score. At no point do we materialize the $N \times N$ correlation matrix or any block of pairwise correlations.

For numerical stability, the normalization uses a small $\epsilon$ in the denominator for near-zero centered vectors. In this case, the implementation subtracts the actual self term $\mathbf{g}_i^\top\mathbf{g}_i$ in the leave-one-out formula; when vectors are exactly unit-normalized, this term reduces to 1. Reductions are accumulated in FP32, and GPU kernels perform the chunk-wise normalization, reductions, and dot products. CPU work is limited to feature loading, chunk scheduling, and writing scores.

For $N$ samples with feature dimension $d$, the score computation costs $\mathcal{O}(Nd)$ after feature extraction. With chunk size $B$, the active memory is $\mathcal{O}(Bd+d+N)$ if all scores are kept in memory, or $\mathcal{O}(Bd+d)$ when scores are streamed to disk. This aggregate streaming implementation explains why processing LLaVA-665K takes only 1.5 hours on an A100 80GB GPU, far below the 94 hours required for full fine-tuning and the 87 hours required by the selection stage of TIVE.

\paragraph{Interpretation of the redundancy score.}
PRISM's redundancy score should be interpreted as a corpus-level average-correlation degree under the centered-cosine kernel, rather than as a local nearest-neighbor density estimator. This design is intentional. Local density methods such as kNN-based scoring can capture fine-grained neighborhoods, but they require either approximate nearest-neighbor search or explicit pairwise comparisons at scale. PRISM instead estimates whether a sample aligns with the dominant centered semantic mass of the corpus. Samples with low average correlation are less explained by this corpus-level semantic mass and are therefore more likely to contribute complementary information during instruction tuning.

\paragraph{Outliers versus useful diversity.}
A potential concern is that low-correlation samples may include outliers. In practice, instruction-tuning datasets are already curated to contain valid image-instruction pairs, and PRISM is applied within this curated pool. We therefore interpret low correlation not as arbitrary outlierness, but as low redundancy relative to the instruction corpus. The qualitative examples in Appendix K further show that high-ranked samples tend to contain more specific visual reasoning, OCR, or compositional information, whereas low-ranked samples are often visually or semantically repetitive.
\end{document}